\documentclass[acmsmall, nonacm]{acmart}
\usepackage[utf8]{inputenc}

\usepackage{graphicx}
\usepackage{subcaption}
\usepackage{xurl}
\usepackage{tabularx,booktabs}
\usepackage{amsmath}

\begin{document}

\title{Epistemic diversity across language models mitigates knowledge collapse}

\author{Damian Hodel and Jevin D. West}

\affiliation{%
  \institution{Center for an Informed Public, Information School, University of Washington}
  \country{USA}
}
\email{hodeld@uw.edu}
\email{jevinw@uw.edu} 
\thanks{ 
This work is supported by the University of Washington’s Center for an Informed Public and the John S. and James L. Knight Foundation (G-2019-58788).
}


\begin{abstract} 
Artificial intelligence (AI) increasingly generates the very content used to train future AI systems. This feedback loop can degrade model quality, reduce informational diversity, and ultimately drive knowledge collapse, i.e. a degradation to a narrow and inaccurate set of ideas. We ask: to mitigate collapse, is it better to concentrate the internet's knowledge into a handful of dominant models (referred to as an AI monoculture), or to distribute it across a diverse ecosystem of models? 
To study the effect of diversity on model performance, we randomly segment the fixed training data across an increasing number of language models and evaluate the resulting \textit{ecosystems} of models over ten self-training iterations.
Our results show that diversity improves long-term performance of models, while monoculture accelerates collapse. 
Specifically, we observe that the \textit{optimal} diversity level (i.e., the level that maximizes performance) increases monotonically with the number of self-training iterations.
The observed effect is robust across various experimental settings, including different model families, parameter sizes, mixing human- and model-generated data, and temperature sampling methods, demonstrating the significance of ecosystem diversity for mitigating collapse.  Moreover, our experiments with increased model and dataset sizes indicate that scaling up the system can amplify collapse in homogeneous ecosystems, thereby increasing the diversity benefits.
In the presence of AI monoculture, our results suggest considering information environments with specialized AI models that maintain and enhance diversity in knowledge production, akin to the benefits of ecological diversity in biology and social systems.\footnote{
The code is available through GitHub at \url{https://github.com/hodeld/diversity-model-collapse}.}.
\end{abstract}

\maketitle

\section{Introduction}

Across social and biological domains, history teaches us the risks of monocultures and the importance of diversity for systemic resilience. Monocropping and the resulting spread of potato blight contributed to the Irish Potato Famine of the 1840s \cite{editors_irish_2017}, while the correlated behavior of large, interconnected banks was a key precondition for the 2007–2008 global financial crisis \cite{butzbach_systemic_2016}, to name just two dramatic examples. Across contexts, increasing diversity has emerged as a successful strategy for mitigating such systemic risks, whether through species diversity in ecosystems \cite{lindenmayer_hidden_2018, otero_biodiversity_2020}, banking diversity in financial market \cite{haldane_systemic_2011, beale_individual_2011}, or even diversity among scientists in knowledge production \cite{zollman_epistemic_2010, bogdan_functional_2026, solomon_norms_2006, nielsen_gender_2017}.

The term \textit{monoculture} has also been adopted by computer scientists to describe the current artificial intelligence (AI) \textit{ecosystem}, in which large generative models trained on vast portions of the internet produce broadly homogeneous outputs both across and within models \cite{lazar_ai_2023, xu_echoes_2025, zhang_cultivating_2025}. As these models are increasingly integrated into scientific writing and discovery \cite{liang_quantifying_2025, hao_artificial_2026}, this homogenization may contribute to the systemic risk of \textit{knowledge collapse}: a degradation toward a narrow and potentially inaccurate set of ideas \cite{peterson_ai_2025, wright_epistemic_2025, messeri_artificial_2024}.
This systemic risk is further amplified by a closely related epistemic failure known as \textit{model collapse}: the degradation in performance and output quality that occurs when models are recursively trained on their own outputs \cite{herel_collapse_2024, shumailov_ai_2024, bertrand_stability_2024}.

Because model collapse is observed at the level of individual models, it is unsurprising that most mitigation strategies, such as output diversification and filtering, have likewise been developed by studying \textit{individual} models \cite{xing_caveats_2025}. However, in real-world settings, \textit{multiple} models evolve collectively through retraining on artificial data \cite{pedreschi_human-ai_2025}.

Motivated by insights from other ecosystems, particularly science as a system of knowledge production, we ask whether increasing ecosystem diversity in AI (i.e., the number of dissimilar models) can mitigate model collapse, and whether any such effect increases monotonically or reaches an optimum.

%

In a homogeneous ecosystem, collapse occurs because of a model's non-zero probability of losing information or learning wrong patterns, which is irreversible, when relying only on its own output \cite{shumailov_ai_2024}. Analogous to productive disagreement among humans, we hypothesize that dissimilar models trained on their collective outputs can improve collective performance by correcting each other's errors. Viewing an AI model as an epistemological representation, we use the Hill-Shannon Diversity (HSD, $D$) from biology \cite{roswell_conceptual_2021, jost_entropy_2006, hill_diversity_1973} to measure ecosystem diversity, which quantifies the effective number of distinct AI models. It is important to note that \textit{the diversity across models serves as the primary independent variable that we manipulate to examine its effect on performance}. This stands in contrast to prior work, which has focused on diversity in data \cite[e.g.,][]{schaffelder_synthetic_2025, zhu_what_2025} and/or has treated diversity as an outcome variable \cite[e.g.,][]{wright_epistemic_2025}.  

Given the enormous resources required to train and operate prevailing models (e.g., ChatGPT, Gemini, Claude, etc.), studying the impact of diversity on AI ecosystem evolution at real-world scale is infeasible. And, even if we had infinite resources, many confounding factors could shape model performance, including model and data size \cite{kaplan_scaling_2020, bertrand_stability_2024}, model architecture, and system design. To address these challenges, we build upon previous work that analyzes a simple, single-model in a semi-controlled environment \cite[e.g.][]{shumailov_ai_2024, herel_collapse_2024}. These prior studies re-train small language models on their own outputs and evaluate performance by measuring perplexity on a fixed ground-truth dataset across multiple (e.g., ten) self-training iterations. We take a similar approach, but instead focus on ecosystems of models (i.e., collections of models), rather than a single model. This allows us to study the effects of ecosystem diversity on model performance, which we measure separately on two distinct model families (OPT \cite{zhang_opt_2022} and GPT2 \cite{radford_language_2019}). 

\begin{figure}[ht]
    \centering
    \includegraphics[width=.8\linewidth]{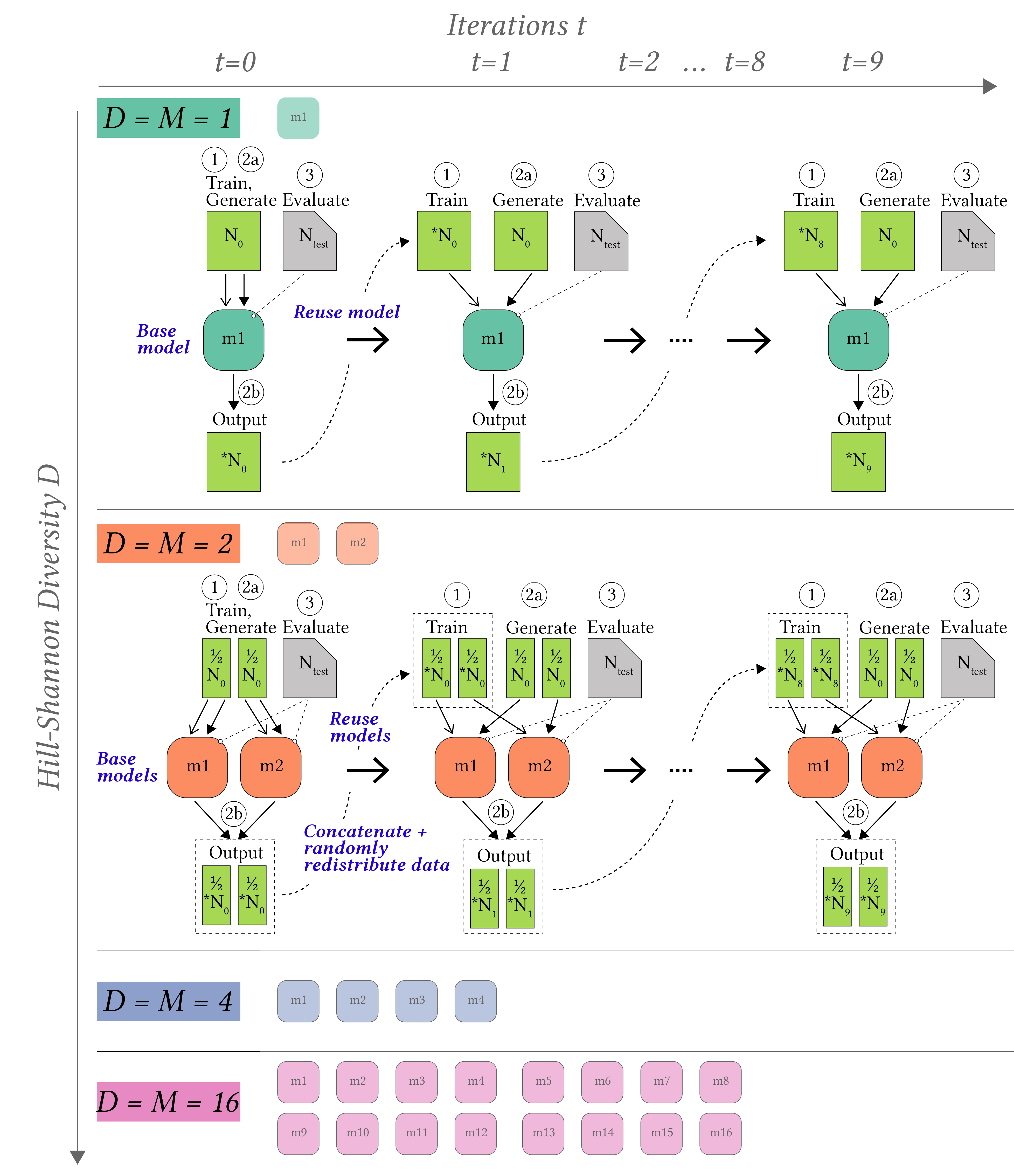}
    \caption{\textbf{Experimental design: manipulating ecosystem diversity and evaluating performance across self-training iterations.} We examine the performance of models $m$ within collections of increasing diversity $D$ when trained on collective output across 10 self-training iterations $t$. The number of models $M$ equals the Hill-Shannon Diversity $D$, which is manipulated through segmentation of the training data. At each iteration, we concatenate and shuffle the $M$ generated datasets, and uniformly redistribute to the $M$ models for the next step. This segmentation approach keeps the data size (i.e., number of tokens) per ecosystem $N$ fixed, yielding a per-model sample size of $n = N/M$. The ground-truth dataset used to measure performance, the original Wikitext2 test split, is fixed across all models and iterations. Thus, each iteration consists of three intermediate steps: Training the model (1, ``Train''), generating outputs using a generation dataset (2, ``Generate''), and evaluating the model (3, ``Evaluate''). In the first iteration step, the training data and the generation dataset are identical for a given model. The models $m$ are identical at step 0 and are continuously fine-tuned on their (collective) outputs. 
    }
    \label{fig:approach}
\end{figure}

While in practice systems can be differentiated in many ways, (e.g., by size, architecture, post-training, system messaging)\footnote{A specific model architecture can be used to build different AI systems (i.e., the end products with which users interact). For simplicity, we use ``model family'' to differentiate between OPT and GPT2 pre-trained base models and ``model'' to refer to the fine-tuned versions.}, \textit{we instantiate ecosystem diversity through variation in training data}. To do this under a controlled setting, we randomly segment the training data, which is fixed in size and, in the first iteration step, also in content. Through this process, we go from one, pre-trained model to multiple fine-tuned models ($M$=1, 2, 4, 16), as shown in Figure~\ref{fig:approach}. This segmentation approach allows us to vary ecosystem diversity ($D=M$) in a relatively simple manner, while keeping factors, such as model family and the per-ecosystem resource budget for sample size and computation, fixed. We use Wikitext2 \cite{merity2016pointer} as the dataset for generating artificial training data, evaluation, and initial training. In our primary experimental setting, we use the smallest versions of the two model families (OPT-125m and GPT2).\\ 
Our experiments identify an optimal level of ecosystem diversity for model performance, which increases monotonically with the number of self-training iterations (Figure~\ref{fig:opt_diversity_iterations}). In other words, the prevailing strategy of training a single model on as much data as possible is short-sighted: over longer horizons, a collection of models trained on segmented datasets outperforms a single model. As the number of self-training loops increases, stronger segmentation is required to reach the diversity level that maximizes overall performance. Put simply, the benefits of diversity grow with the number of iterations.

To assess the effect of ecosystem diversity under more realistic conditions, we introduce two variations of our primary setting.
In the first variation (V1), we increase the size of system (i.e., model and data size). Scaling up leads to a stronger collapse in more homogeneous ecosystems, while collapse is attenuated in more diverse settings, resulting in increased diversity benefits. 
The second variation (V2) enhances the training data by (1) incorporating 10\% fresh, original Wikitext2 data at each iteration and (2) by applying temperature sampling with $\tau = 0.5, 1, 2$. Prior work suggests that both real-data incorporation \cite{shumailov_ai_2024, gerstgrasser_is_2024, xing_caveats_2025} and data diversification \cite{wright_epistemic_2025, schaffelder_synthetic_2025, zhu_what_2025} can mitigate collapse. While our results confirm these findings, \textit{they also show that the effect of ecosystem diversity outweighs the effects of data quality manipulation}.
Overall, the experimental variations suggest that the benefits of diversity for long-term performance may persist under real-world AI ecosystem conditions, where models such as ChatGPT are substantially larger, trained on far more data, can be optimized for diverse outputs, and where human-generated data is not replaced but instead accumulates alongside artificial data.


As AI models become increasingly incorporated into major institutions such as science, education, government, and economics, a collapse toward the most dominant ideas could have severe societal implications.
The results of our experiments suggest that differentiation of models could be a significant factor in mitigating such collapse. 

\section{Approach}
We study the progress of model performance in AI ecosystems containing 1, 2, 4, and 16 models, each re-trained on their collective output over 10 iterations. All models begin as identical copies of a pre-trained model but are fine-tuned on equal-sized, non-overlapping subsets of the training data\footnote{This approach using non-overlapping train sets is sometimes called bootstrap aggregation (bagging) \cite{ganaie_ensemble_2022} and has been applied in early work studying performance of model ecosystems \cite[e.g.,][]{krogh_neural_1994}}. This segmentation approach enables us to isolate and examine the effect of diversity across models on their performance under fixed model architecture and a fixed per-ecosystem resource budget for data size (number of tokens, $N$) and compute. 
We first introduce the primary experimental setting, which demonstrates a clear positive effect of ecosystem diversity on model performance in a controlled, though limited, setting. Then, in Section~\ref{sec:approach_variations}, we present variations of this primary setting that aim to approximate more realistic conditions and to examine how the location of the optimum shifts with factors such as model size, data size, incorporation of real data into training, and temperature sampling.
We denote the following main variables: $D$ (ecosystem diversity), $M$ (number of models per ecosystem), $N$ (fixed data size per ecosystem, i.e., number of tokens), $n$ (data size per model), $T$ (number of iterations $t$, i.e., 10), and $\mu$ (model perplexity).

Each iteration step involves an evolving \textbf{training dataset} used to update model weights, a fixed \textbf{generation dataset} used to generate artificial training data for the next iteration, and a fixed \textbf{evaluation dataset} used to measure model perplexity (see Figure~\ref{fig:approach}). The evaluation dataset is the same for all models and corresponds to the original Wikitext2 test set. The generation datasets are unique to each model but fixed across all iterations, corresponding to randomly sampled subsets of the original Wikitext2 training data.

In the first iteration step, the training data is identical to the generation dataset for a given model. Thereafter, it evolves in content because it is sourced from the outputs of the previous iteration: At each iteration, the $M$ generated datasets are concatenated, shuffled, and uniformly redistributed to the $M$ models for the next round. Throughout the ten iterations, the training and generation datasets for each model within an ecosystem are equal in size, $n = N/M$. For example, for two models ($D;M = 2$), the sizes of the training and generation datasets are each half of the original size, $n = N/2$ (see Figure~\ref{fig:approach}).
In the case of the single-model ecosystem ($D;M=1$), our approach parallels the methods used in prior work on single-model collapse \cite[e.g.,][]{shumailov_ai_2024}.


Translated to the real world, the segmentation approach broadly mirrors domain- or community-specific models \cite[e.g.,][]{tanksley_ethics_2025, costanza-chock_design_2020} trained on distinct segments of internet data. While tailored to different communities that might use different inputs, the generated outputs spread across the internet and later become part of the training data for future models.  

\subsection{Ecosystem diversity D and model performance}
Drawing from ecology, we use the HSD to measure ecosystem diversity, that is, diversity across models. The HSD quantifies the effective number of distinct, equally frequent models in a ecosystem \cite{roswell_conceptual_2021, jost_entropy_2006, hill_diversity_1973}. Let $M$ denote the number of models, then we calculate HSD as:

\begin{equation}\label{eq:hill_shannon}
D(M) = \exp\!\left( -\sum_{m=1}^{M} w_m \ln w_m \right)
\end{equation}

Because of equal weight across models ($w_m=1/M$), the diversity simplifies to the number of models in a given ecosystem (i.e. richness):

\begin{equation}\label{eq:hill_shannon_simple}
D(M) = M
\end{equation}

Consistent with prior work \cite{shumailov_ai_2024}, we compute each model's perplexity on the original Wikitext2 test split to assess model performance. Perplexity reflects a model's ability to complete the test set text sequences (lower is better) and indicates the degree of alignment between the model's probability distribution $p_{\theta}$ and the true distribution $p_{data}$. 

In an AI ecosystem composed of $M$ different models, we calculate the combined mean perplexity $\mu$ for a given iteration step $t$ as:

\begin{equation}\label{eq:p_eco}
    \mu_t = \frac{1}{M} \sum_{m=1}^{M} \mu_{m, t} \text{ ,  } 
\end{equation}
whereas $\mu_{m, t}$ corresponds to the mean perplexity across the text sequences of the original Wikitext2 test split.  

To compare ecosystems across iterations, we calculate an aggregated performance. 
Let $\mu_t$ an ecosystem's mean perplexity for a given iteration step $t$, then, the aggregated mean perplexity $\mu_T$ across $T$ iterations equals: 

\begin{equation}\label{eq:p_macro}
    \mu_{T} = \frac{1}{T} \sum_{t=1}^{T} \mu_t
\end{equation}

Finally, for a fixed number of iterations $T$ and a set of ecosystems with varying diversity levels, we define the optimal diversity level as the one that achieves the highest aggregated performance:
\begin{equation}\label{eq:optimum_level}
D_{\mathrm{opt}}(T) = \arg\min_D \mu_T(D).
\end{equation}


\subsection{Generation and training details}
We use OPT \cite{zhang_opt_2022} and GPT2 \cite{radford_language_2019}, two open-accessible, small, and well-studied language model families used in prior work on collapse \cite[e.g.,][]{shumailov_ai_2024, herel_collapse_2024}. For the primary approach, we use the smallest versions, OPT-125m and GPT2, with 125M and 124M parameters, respectively. 


Analogous to prior work \cite{herel_collapse_2024, shumailov_ai_2024}, we use Wikitext2 \cite{merity2016pointer} as the original dataset. To reduce computational effort during training and generation, the primary approach uses a randomly sampled subset of two-fifths of the original size per ecosystem ($N =0.8$M tokens), while keeping the original Wikitext2 validation and test splits for evaluation. Using a block size of 128 tokens, the sizes of the training–generation, validation, and test splits correspond to 7,500, 1,900, and 2,200 data points, respectively.

Our approach for generating the artificial dataset follows a method introduced in prior work \cite{shumailov_ai_2024}. For each available token sequence of length 128 in the generation dataset, we prompt the model to generate a continuation sequence of equal length, using a five-way beam search. Because we iterate through the entire generation dataset, the model would produce the original generation dataset if it had zero error. 
At each iteration, each model is trained for 5 epochs with a learning rate of $1e^{-5}$ on a training sample of size $n=N/M$, using a block size of 128 tokens. The best-performing model on the validation set is preserved for subsequent iterations. 

To study the effect of increasing parameter size, we use OPT-350M with 350M parameters and the full Wikitext2 dataset (2.1M tokens); see Section~\ref{sec:approach_variations}. The analogous experiments with GPT2-Medium (355M parameters) are reported in Appendix~\ref{sec:gpt2_scaling}.
All experiments are carried out using the HuggingFace Transformers library \cite{wolf2019huggingface} and the Python Lightning framework \cite{Falcon_PyTorch_Lightning_2019}. 

\subsection{Variations of primary setting} 
\label{sec:approach_variations}
To understand the effect of diversity in more realistic settings, we examine how the location of the optimum shifts under two types of experimental variations: V1 examines the effect of increasing model and data sizes, and V2 examines the effect of data quality improvements through real-data incorporation and temperature sampling. We conduct the experimental variations using the OPT model family and report the scaling experiments with GPT2 in Appendix~\ref{sec:gpt2_scaling}. 

\paragraph*{V1: System scale –– model size and data size} 
In reality, both model and data sizes are far larger. OPT-125m contains 125M parameters, whereas current large language models (e.g., OpenAI's ChatGPT, Google's Gemini, Anthropic's Claude) likely contain in the order of trillions of parameters. A similar discrepancy applies to dataset size: while Wikitext2 contains roughly 2M tokens, the datasets used for prevailing models are likely $10^{6}$ times larger.
Established scaling laws suggest a power-law relationship between performance and parameter size \cite{kaplan_scaling_2020, hoffmann_training_2022}. However, recent work indicates that these relationships may break down under self-training conditions and that larger models and datasets may even amplify collapse \cite{dohmatob_strong_2024, dohmatob_tale_2024}.
To examine ecosystem diversity through the lens of scaling laws we introduce three variations of the primary setting: (1) a larger model (OPT-350m, with 350M parameters), (2) the full Wikitext2 training set (2.1M tokens), and (3) the combination of both larger model and dataset sizes. We conduct these experiments for $M=$1, 2, 4, 8, and 16 models. 

\paragraph*{V2: Training data quality –– real-data incorporation and sampling temperature}
In realistic self-training loops, artificial data does not replace but rather accumulates alongside real (non-model-generated) data, improving the quality of the training set and mitigating collapse \cite{shumailov_ai_2024, gerstgrasser_is_2024, xing_caveats_2025}. Similarly, prevailing language models are not deterministic but incorporate randomness that diversifies outputs and can mitigate model collapse \cite{wright_epistemic_2025, schaffelder_synthetic_2025, zhu_what_2025}. A common mechanism is temperature sampling, which increases output diversity by flattening token-sampling probabilities \cite{chung_increasing_2023, alihosseini_jointly_2019}.
To test these factors, we introduce an experimental variation that replaces 10\% of the model-generated data with randomly sampled original Wikitext2 training data at each iteration (following \cite{shumailov_ai_2024}), as well as three additional variations using sampling temperatures of $\tau = 0.5$, $1.0$, and $2.0$. As a reminder, the primary experiment uses deterministic generation with $\tau = 0$. When temperature sampling is applied, all remaining sampling parameters follow the Transformers default settings, corresponding to \texttt{top\_k} = 50 and \texttt{top\_p} = 1.0.

\section{Results}
We conducted a series of experiments to examine how diversity across models affects performance when models are trained on their collective outputs. Under a fixed data size and compute budget, we find that the optimal degree of diversity increases monotonically with the number of self-training iterations:
\begin{equation}\label{eq:optimum_level_monotonic}
T_1 < T_2 \Longrightarrow D_{opt}(T_1) \leq D_{opt}(T_2), 
\end{equation}
where the optimal level is defined as the ecosystem achieving the highest aggregated performance (i.e., lowest aggregated mean perplexity). 
In other words, as the number of self-training iterations increases, maximizing overall performance requires segmenting the data across more models. Correspondingly, the benefits of diversity relative to the most homogeneous setting grow with the number of iterations.

\begin{figure}[ht]
    \centering
    \includegraphics[width=0.9\linewidth]{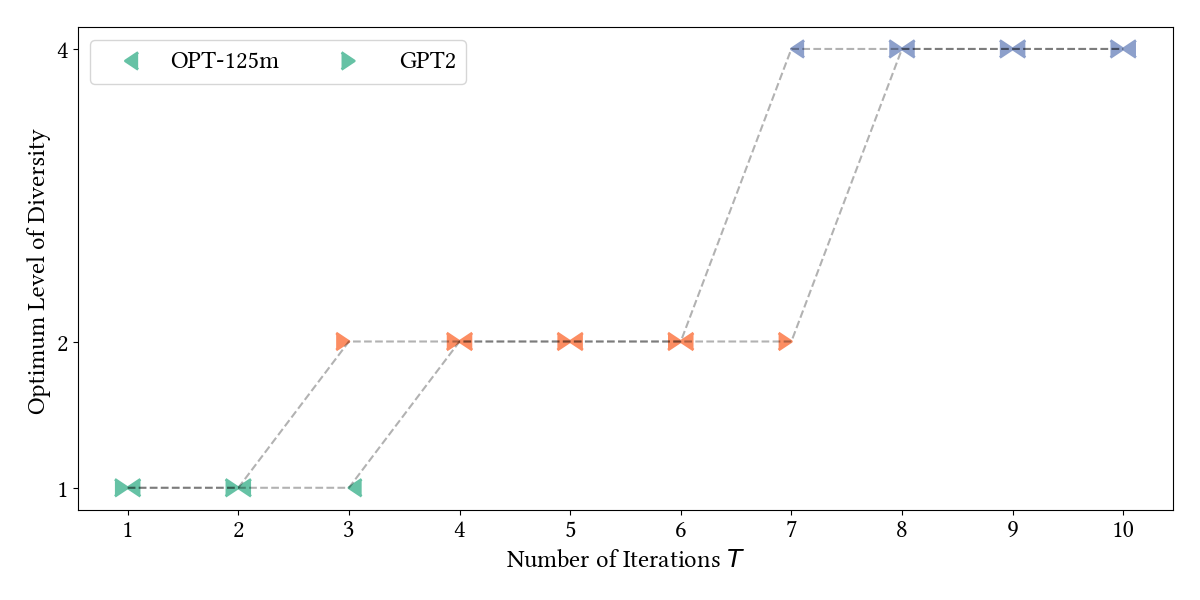}
    \caption{\textbf{The optimal level of ecosystem diversity (Hill–Shannon diversity $D$) increases monotonically with the number of self-training iterations $T$.}
    The most homogeneous setting ($D=1$, green) performs best only in the short term ($T \leq 3$ for OPT-125M and $T \leq 2$ for GPT2). Thus, diversity $D$ mitigates model collapse. As $T$ grows, the data must be segmented across models to reach the diversity level that maximizes overall performance on the original test set ($D_{opt}(T) = {\mathrm{argmin}_D} \, \mu_T(D)$). 
    The corresponding performance trajectories are presented in Figure~\ref{fig:both_evolution}. Colored symbols correspond to the diversity levels shown in Figure~\ref{fig:approach} 1, 2, 4, and 16 (e.g., orange denotes $D=2$). Intermediate levels are omitted, as the key result is the monotonic increase in diversity benefits with $T$. 
    An extended range of up to 20 iterations is shown in Figure~\ref{fig:opt_20} in the Appendix.
    }
    \label{fig:opt_diversity_iterations} 
\end{figure}

\begin{figure}[htbp]
    \centering
    \begin{subfigure}[b]{0.75\linewidth} 
        \centering
        \includegraphics[width=\linewidth]{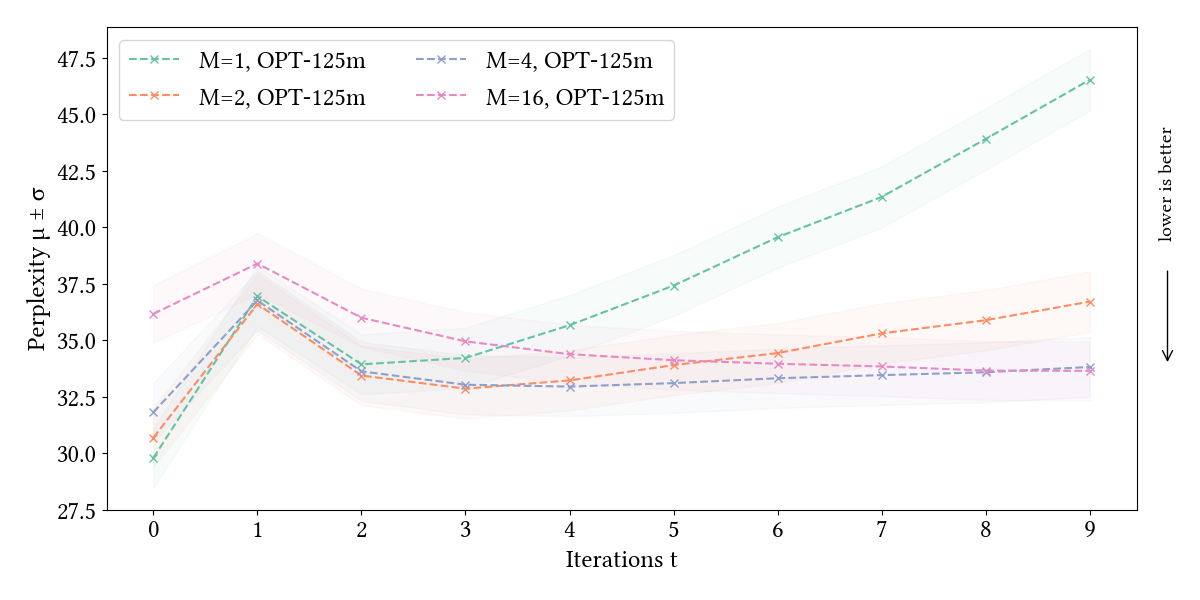}
        \caption{OPT-125m}
        \label{fig:opt_slice}
    \end{subfigure}
    \\ 
    \begin{subfigure}[b]{0.75\linewidth}
        \centering
        \includegraphics[width=\linewidth]{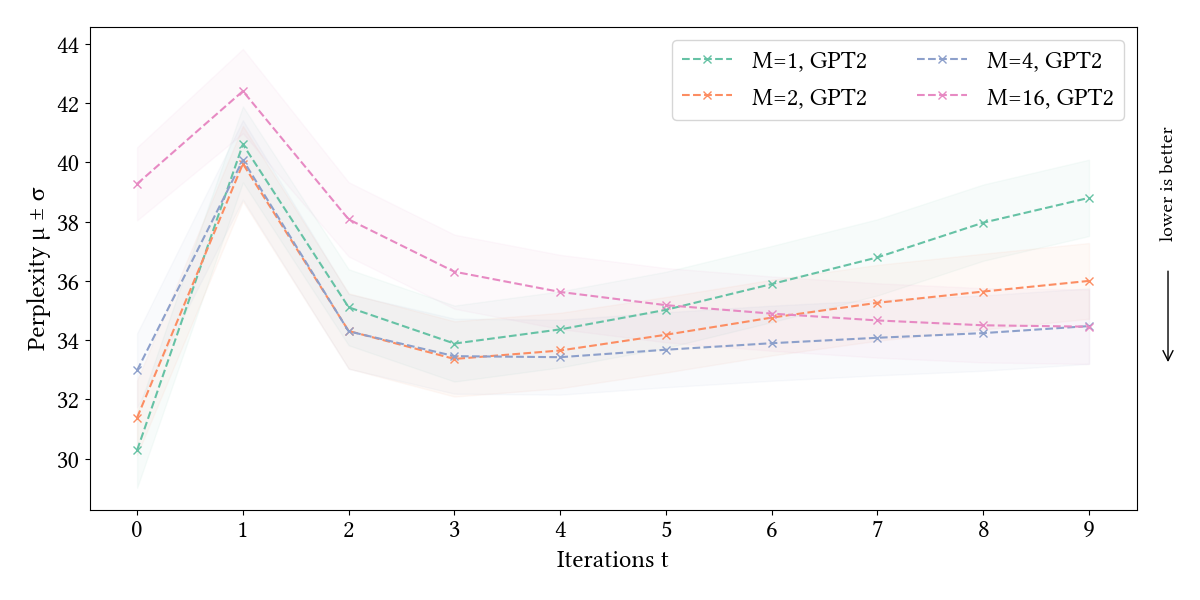}
        \caption{GPT2}
        \label{fig:gpt2_slice}
    \end{subfigure}
    \\ 
    \begin{subfigure}[b]{0.75\linewidth}
        \centering
        \includegraphics[width=\linewidth]{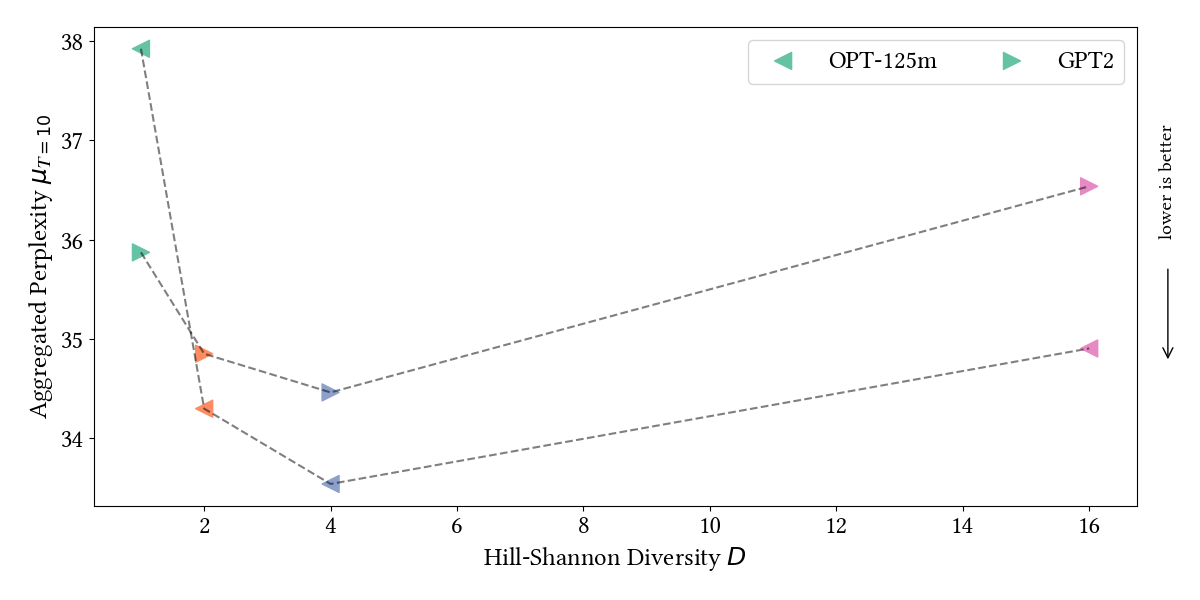}
        \caption{Aggregated perplexity after 10 iterations}
        \label{fig:macro_perplexity_vs_diversity}
    \end{subfigure}
    
    \caption{\textbf{Ecosystem diversity mitigates collapse.}  (a, b) Comparison of ecosystems' mean perplexity with increasing diversity ($D=M$) at fixed dataset size across ten iteration steps. While low-diversity ecosystems ($D=1,2$) exhibit strong performance decay, the high-diversity ecosystem ($D=16$) starts with relatively high perplexity but improves over $t$ . 
    The shaded region is a combined standard deviation across the models and data points. 
    (c) Aggregated perplexity. Both model families exhibit an optimal level of diversity of 4 for overall performance across 10 iterations.
    Colors correspond to the diversity levels shown in Figure~\ref{fig:approach} (e.g., orange denotes $M;D=2$).}
    \label{fig:both_evolution}
\end{figure}

In the primary approach, the single-model setting ($D=1$) performs best only in the short term (considering fewer than four iterations for OPT-125M and fewer than three iterations for GPT2,  see Figure~\ref{fig:opt_diversity_iterations})
Figure~\ref{fig:both_evolution} shows the corresponding mean perplexity trajectories for ecosystems consisting of 1, 2, 4, and 16 models trained on equal-sized, non-overlapping segments of model-generated data across 10 iterations. Ecosystems with lower diversity ($D;M=1,2$) perform well initially but exhibit rapidly increasing perplexity as the number of iterations grows. In contrast, the most diverse ecosystem ($M=16$) starts with relatively high perplexity but improves over time. Across the full 10 iterations, a diversity level of four ($D;M=4$) yields the lowest aggregated mean perplexity for both model families, with a stronger effect for OPT-125M. The least diverse ecosystem ($D=1$) performs best only at $t=0$, when ecosystems differ solely in sample size per model ($n=N/M$), as all models are trained on the original data\footnote{The perplexity progression, including the perplexity of the original models, is shown in Figure~\ref{fig:both_evolution_start-1} in the appendix.}.  

Figure~\ref{fig:opt_20} in the Appendix extends the analysis to 20 iterations for OPT-125M, showing that the optimal diversity level continues to increase beyond 10 iterations.



\begin{figure}[ht]
    \centering
    \begin{subfigure}[t]{0.49\textwidth}
        \centering
        \includegraphics[width=\linewidth]{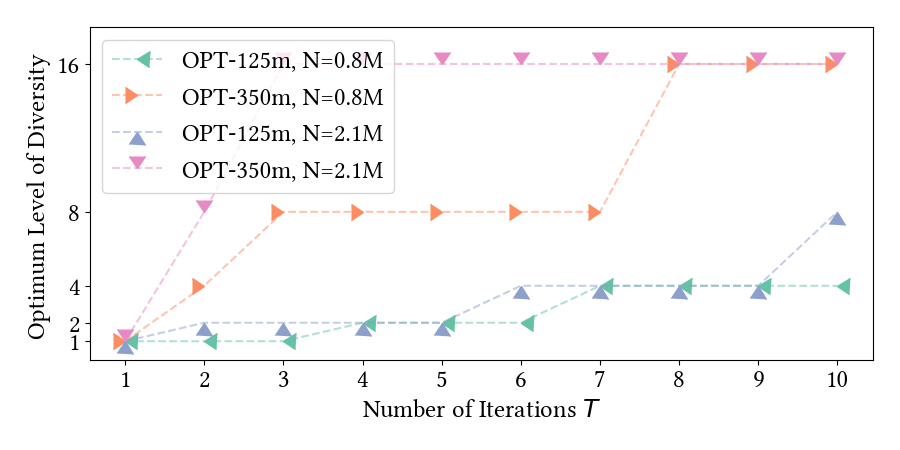}
        \caption{Optimal diversity levels.}
        \label{fig:opt_optimumdiversity_scaling}
    \end{subfigure}
    ~ 
    \begin{subfigure}[t]{0.49\textwidth}
        \centering
        \includegraphics[width=\linewidth]{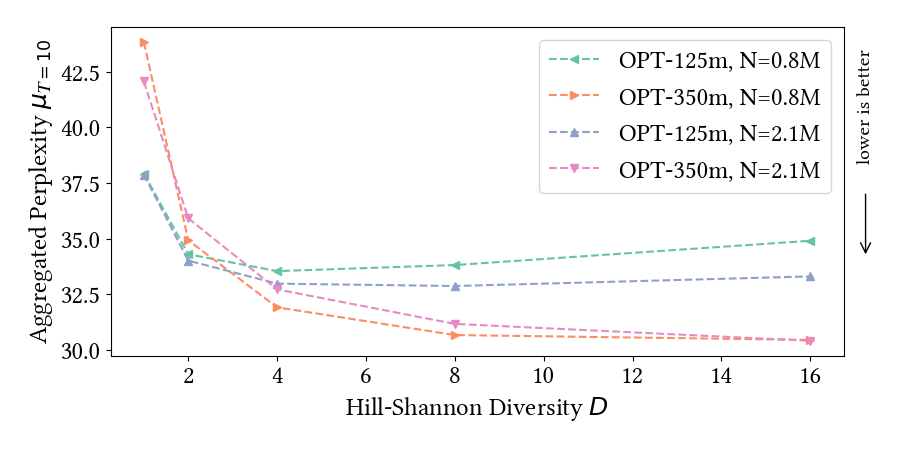}
        \caption{Aggregated perplexities after 10 iterations. 
        }
        \label{fig:opt_agg_pp_scaling}
    \end{subfigure}
    \\ 
    \centering
    \begin{subfigure}[t]{0.49\textwidth}
        \centering
        \includegraphics[width=\linewidth]{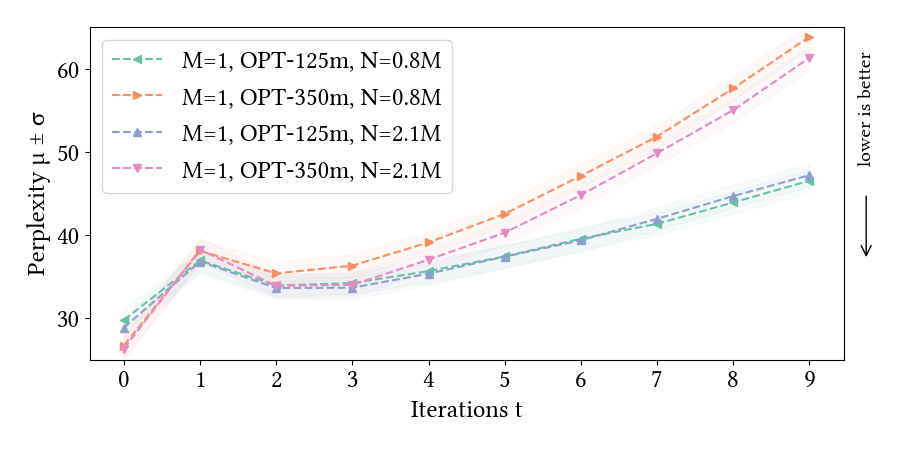}
        \caption{Mean perplexities for M;D=1}
        \label{fig:opt_scaling_m1}
    \end{subfigure}
    ~ 
    \begin{subfigure}[t]{0.49\textwidth}
        \centering
        \includegraphics[width=\linewidth]{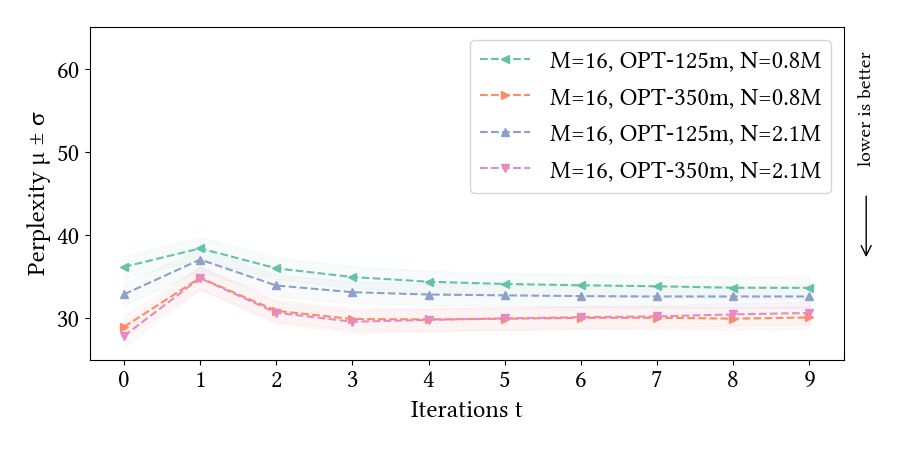}
        \caption{Mean perplexities for M;D=16}
        \label{fig:opt_scaling_m16}
    \end{subfigure}
\caption{
\textbf{The optimal level of diversity increase with model and dataset scale.}
Ecosystem diversity across four system scales (two model sizes × two dataset sizes; variation V1). ``OPT-125M, $N=0.8$M'' (green $\blacktriangleleft$) corresponds to the primary setting, using a 125M-parameter model and a total training dataset of $0.8$M tokens. 
\textbf{Top:} (a) Optimal diversity level vs.\ number of iterations $T$ and (b) aggregated perplexity $\mu_{T=10}$ for five diversity levels ($M;D=1,2,4,8,16$). Larger models (orange and pink vs.\ green and blue) show a steeper increase in optimal diversity with $T$, shifting the optimum from $D=4$ to $D=16$ after 10 iterations. Increasing dataset size (pink $\blacktriangledown$ and blue $\blacktriangle$ vs.\ green and orange) mainly affects larger models, increasing the optimal diversity level when considering fewer than 8 iterations.
\textbf{Bottom:} Perplexity trajectories across ten iterations for two diversity levels (c: $D=1$, d: $D=16$). For a fixed diversity level, increasing model or dataset size reduces perplexity at $t=0$.
(c) In the most homogeneous ecosystem ($D=1$), larger models show strong perplexity increases in later iterations indicating stronger collapse, while solely increasing dataset size has only a modest effect.
(d) In highly diverse ecosystems ($D=16$), larger models achieve lower perplexity across all iterations, while increasing dataset size benefits primarily the smaller model. 
The shaded region represents the combined standard deviation across models and data points.}
\label{fig:opt_slice_scaling}
\end{figure}

\begin{figure*}[ht]
    \centering
    \begin{subfigure}[t]{0.3\textwidth}
        \centering
        \includegraphics[width=\linewidth]{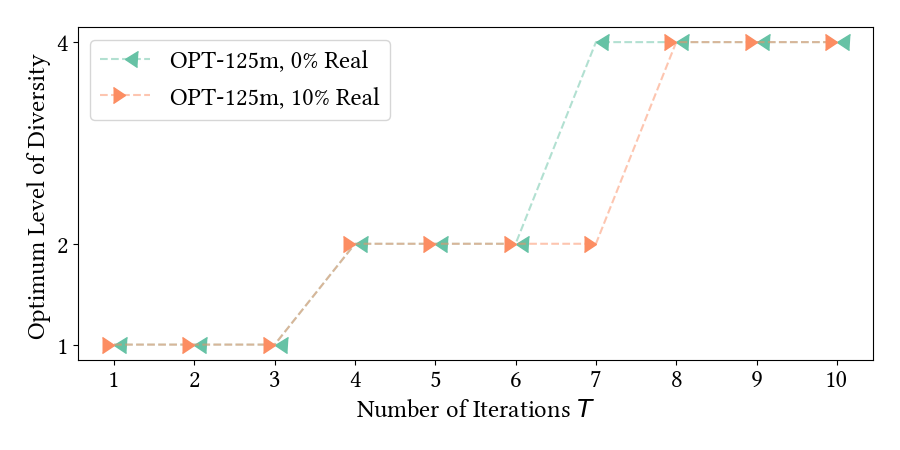}
        \caption{Optimal diversity levels.}
        \label{fig:opt_optimumdiversity_real}
    \end{subfigure}
    ~ 
    \centering
    \begin{subfigure}[t]{0.3\textwidth}
        \centering
        \includegraphics[width=\linewidth]{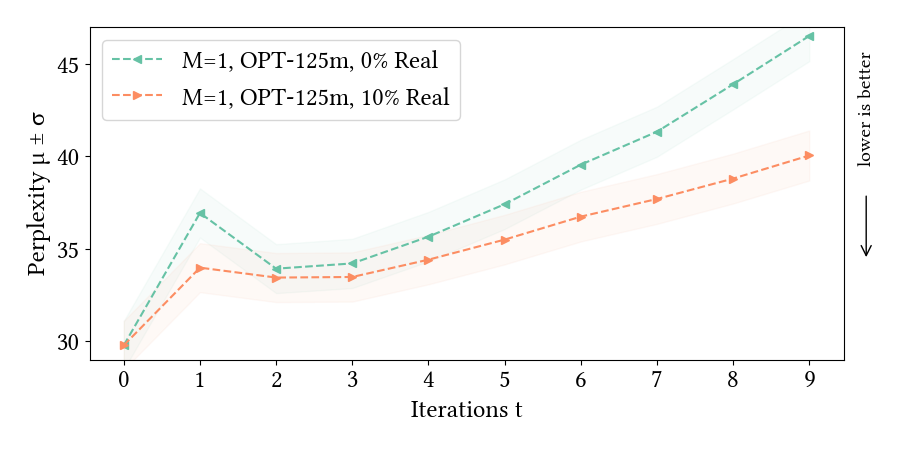}
        \caption{M;D=1}
        \label{fig:opt_real_m1}
    \end{subfigure}
    ~ 
    \begin{subfigure}[t]{0.3\textwidth}
        \centering
        \includegraphics[width=\linewidth]{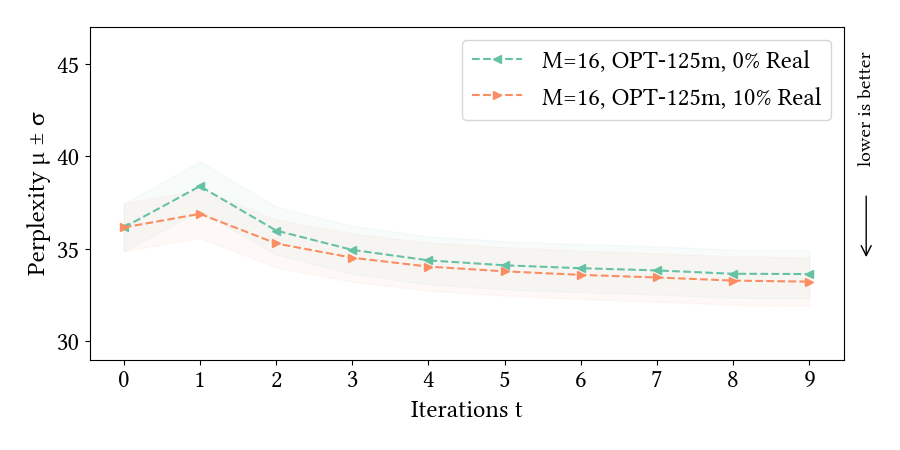}
        \caption{M;D=4}
        \label{fig:opt_real_m16}
    \end{subfigure}
    \caption{\textbf{Replacing 10\% of artificial data with real data has only a minor effect on the optimal diversity level}. 0\% real data (green $\blacktriangleleft$) corresponds to the primary setting.  
    (a) Optimal diversity level vs.\ number of iterations $T$.
    (b, c) Perplexity trajectories across ten iterations for two diversity levels (b: $D=1$, c: $D=16$).
    For a fixed diversity level, replacing artificial data with real data at each iteration reduces perplexity, while differences across ecosystems diminish as diversity increases.}
    \label{fig:opt_slice_realdata}
\end{figure*}

\begin{figure*}[ht]
    \centering
    \begin{subfigure}[t]{0.3\textwidth}
        \centering
        \includegraphics[width=\linewidth]{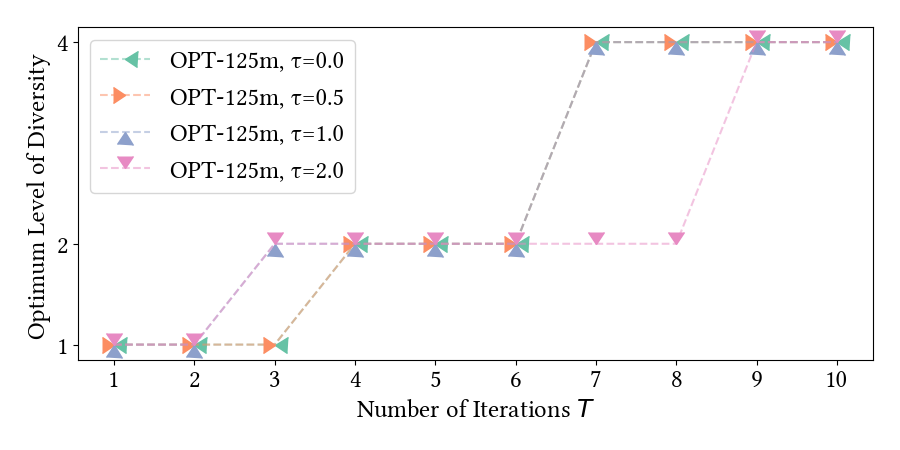}
        \caption{Optimal diversity levels.}        \label{fig:opt_optimumdiversity_temperature}
    \end{subfigure}
    ~ 
    \centering
    \begin{subfigure}[t]{0.3\textwidth}
        \centering
        \includegraphics[width=\linewidth]{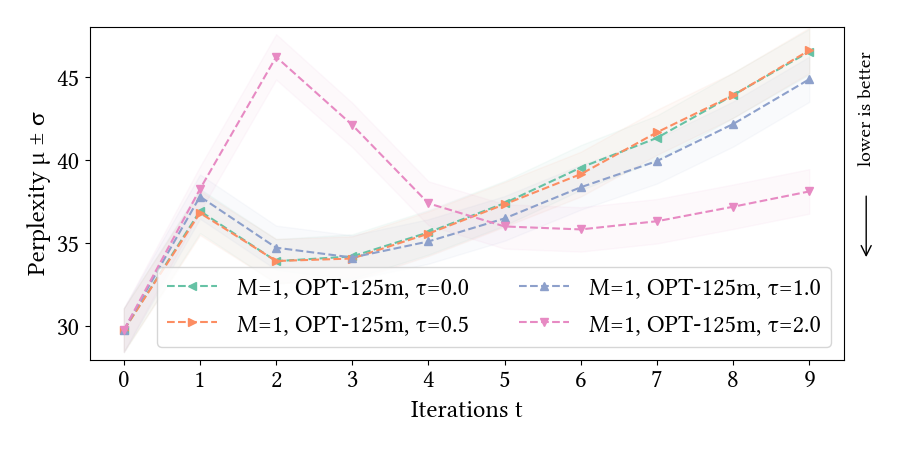}
        \caption{M;D=1}
        \label{fig:opt_temperature_m1}
    \end{subfigure}
    ~ 
    \begin{subfigure}[t]{0.3\textwidth}
        \centering
        \includegraphics[width=\linewidth]{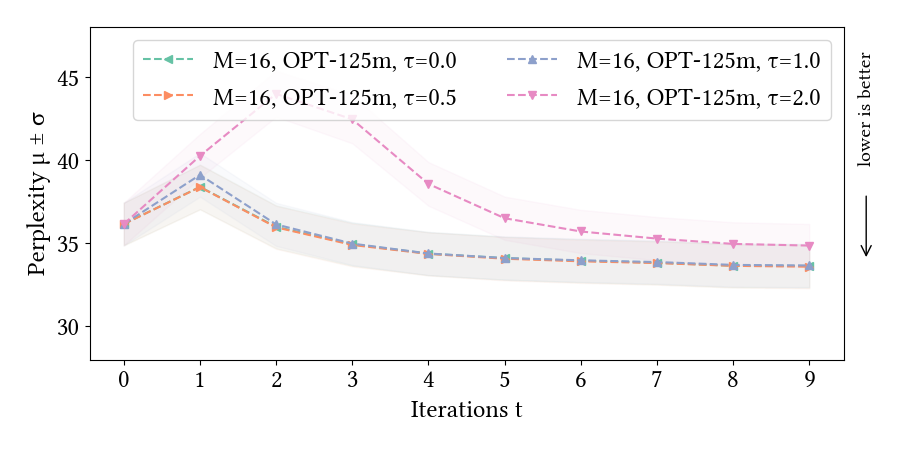}
        \caption{M;D=16}
        \label{fig:opt_temperature_m16}
    \end{subfigure}
    \caption{\textbf{Increasing sampling temperature ($\tau = 0.0, 0.5, 1.0, 2.0$) has only a minor effect on the optimal diversity level.}
    $\tau = 0.0$ (green $\blacktriangleleft$) corresponds to the primary setting.
    (a) Optimal diversity level vs.\ number of iterations $T$.
    (b, c) Perplexity trajectories across ten iterations for two diversity levels (b: $D=1$, c: $D=16$).
    A sampling temperature of $\tau = 1.0$ (blue $\blacktriangle$) slightly reduces perplexity in the most homogeneous ecosystem ($D=1$) and has almost no effect in the most diverse setting ($D=16$). For both diversity levels, $\tau = 2.0$ (pink $\blacktriangledown$) causes a sharp perplexity increase in early iterations, with a relative decrease at later iterations only for $D=1$.}
    \label{fig:opt_slice_temperature}
\end{figure*}

\subsection{V1: Model size and data size} 
\label{sec:result_scaling}

The benefits of diversity increase as both model size (from 125M to 350M parameters) and dataset size (from 0.8M to 2.1M tokens) scale (Figure~\ref{fig:opt_slice_scaling}). Holding dataset size constant, larger models exhibit a steeper increase in the optimal diversity level as the number of iterations grows. A similar pattern appears when increasing dataset size for the larger model, whereas for the smaller model dataset size has only a minor effect on the optimal diversity level. Below, we highlight several notable performance effects relative to the primary setting (125M-parameter model and 0.8M tokens dataset).

At iteration step $t=0$, when models are trained on the original dataset, increasing either model or dataset size improves performance in line with established scaling laws \cite{kaplan_scaling_2020, hoffmann_training_2022}. The most pronounced effects of larger models is the amplified collapse of low-diversity ecosystems ($D \leq 2$), leading to the observed shift in the optimal diversity level.
While the effects of dataset size are less pronounced, two patterns emerge: increasing dataset size for the larger model improves performance primarily in the most homogeneous ecosystem ($D=1$),  whereas for the smaller model it improves performance primarily in the most diverse setting ($D=16$).

\subsection{V2: Real-data incorporation and sampling temperature}
\label{sec:result_data_quality}
Improving data quality by incorporating 10\% real data (Figure~\ref{fig:opt_slice_realdata}) or by applying temperature sampling with $\tau = 1.0$ and $\tau = 2.0$ (Figure~\ref{fig:opt_slice_temperature}) generally improves model performance. Because both interventions primarily benefit the most homogeneous ecosystem ($D=1$), performance differences across diversity levels shrink relative to the primary setting. However, since the positive effect of ecosystem diversity exceeds these improvements, both interventions produce a similar trajectory of the optimal diversity level over iterations as observed in the primary setting. We highlight several notable performance effects relative to the primary setting below.

The impact of incorporating real data at each iteration increases as diversity decreases, reflecting the larger per-model dataset size ($n = N/M$). Temperature sampling can also affect data quality by diversifying generated outputs. For temperatures of $\tau = 1.0$ and $\tau = 2.0$ in the most homogeneous ecosystem ($D=1$), perplexity increases during early iterations but declines by a similar magnitude later in the loop, resulting in only marginal overall improvement. In more diverse ecosystems, $\tau = 1.0$ produces slightly negative effects, while $\tau = 2.0$ leads to strongly negative effects. By contrast, a temperature of $\tau = 0.5$ shows almost no effect relative to the primary setting for any fixed diversity level.

\section{Theoretical interpretation}
\label{sec:theory}
Across experiments, our study of ecosystem evolution reveals a monotonic increase of the optimal diversity level with number of self-training iterations $T$. For a fixed number of 10 iterations, we find a U-shaped relationship between diversity and collapse. Considering that dataset size per model $n=N/M$ decreases as diversity increases, the observed increase in optimum diversity level indicates a complex interplay between opposing effects of diversity and scale with $T$. The complexity grows when considering the results obtained with increased scale (Section~\ref{sec:result_scaling}): When model size increases, we observe stronger collapse in low-diversity ecosystems ($D \leq 2$) and weaker collapse in high-diversity ecosystems ($D \geq 4$). Increasing dataset size generally improves performance for OPT-125m but worsens performance for OPT-350m, except in the $D=1$ ecosystem. In this section we propose a theoretical explanation of the underlying effects.


Dohmatob et al. \cite{dohmatob_strong_2024} theorize that increasing model and/or dataset size amplifies model collapse once data quality becomes sufficiently low. Building on this theory, we can assume that low-diversity ecosystems ($D \leq 2$) tend to operate in a low data-quality regime, leading to stronger collapse as model size increases, while the opposite holds for high-diversity ecosystems ($D \geq 4$). This raises the question of how diversity helps preserve high data quality, given that diverse ecosystems start with lower model performance at iteration step $t=0$.
To address this, we introduce \textit{effective data quality (EDQ)}, which builds on the framework of \cite{dohmatob_strong_2024}. We then use this hypothesis to explain our empirical results in detail.

\subsection*{Effective data quality EDQ}  
Training data quality reflects the discrepancy between $\hat{p}{data}$ and $p{data}$ \cite{dohmatob_strong_2024, wang_theoretical_2025}, for example as measured by \cite{shumailov_ai_2024, bertrand_stability_2024}:
$\delta(\hat{p}_{data}) = \mathbb{W}_2(\hat{p}_{data}, p_{data})$, whereas $\mathbb{W}_2$ refers to the Wasserstein-2 distance between the two distributions\footnote{The probability distributions are illustrated in Figure~\ref{fig:distributions} in the appendix.}.
For an original (non–model-generated) dataset, all examples lie within the support of the true distribution, and this discrepancy approaches zero as dataset size approaches infinity. In contrast, due to models' internal errors (functional expressivity and functional approximation errors \cite{shumailov_ai_2024}), resampled artificial datasets can contain less information about the true distribution and can even include examples outside its support. Moreover, depending on the data received in previous training steps, not every in-distribution example provides new or useful information to a model \cite{feng_beyond_2024, sorscher_beyond_2023} and further  training on a dataset similar to the initial training set risks performance decay via overfitting \cite{tirumala_memorization_2022}. 
Effective fine-tuning methods suggest that optimal training data is not merely data that reflects the true distribution, but data that most efficiently moves model weights toward that distribution \cite{ouyang_training_2022, liu_rethinking_2025, deb_fishersft_2025, antonello_selecting_2021}. Hence, optimal data must account for the model's existing distribution, which approximates a mixture of previous training distributions, i.e., ${p}_{\theta(t)} \approx \sum_{i=0}^{t-1} \hat{p}_{data(i)}$. 
We define EDQ at iteration $t$ as the difference between the training–truth discrepancy and the model–truth discrepancy\footnote{Because we introduce EDQ primarily to explain our empirical results, we rely on a simple scalar metric to capture distributional discrepancy and ignore additional factors required for optimal model weights updates, such as the direction of the distributional shift.}:



\begin{equation}\label{eq:eff_data_quality} 
\delta_{EDQ}(\hat{p}_{data}, {p}_{\theta)} := \mathbb{W}_2(\hat{p}_{data(t)}, p_{data})- \mathbb{W}_2({p}_{\theta(t)}, p_{data}),
\end{equation}

where lower values of $\delta_{EDQ}$ indicate higher EDQ. EDQ reflects the model-dependent informativeness of the given training data.
Using $\delta_{EDQ}$, we can now state our main hypothesis  under finite training data and compute: 

\textbf{Hypothesis: EDQ, informal}

\begin{enumerate}
\item \textbf{High-EDQ regime:}   If $\mathbb{W}_2(\hat{p}_{data(t)}, p_{data}))$ sufficiently lower than $\mathbb{W}_2({p}_{\theta(t)}, p_{data})$, additional training improves performance.

\item \textbf{Low-EDQ regime :}   If $\mathbb{W}_2(\hat{p}_{data(t)}, p_{data}))$ sufficiently greater than $\mathbb{W}_2({p}_{\theta(t)}, p_{data})$, additional training degrades performance.

\item \textbf{Medium-EDQ regime:} If $\mathbb{W}_2(\hat{p}_{data(t)}, p_{data})) \approx \mathbb{W}_2({p}_{\theta(t)}, p_{data})$, additional training neither significantly improves nor degrades model performance.
\end{enumerate}

This hypothesis has two implications for recursively trained models.
First, performance gains or losses depend on EDQ, regardless of whether the data is generated by humans or models \cite{wang_theoretical_2025, feng_beyond_2024, gambetta_learning_2025}, or on its diversity per se. Nevertheless, both the proportion of artificial data and data diversity can shape the EDQ.
Second, when operating in a low-EDQ regime, neural scaling laws invert: performance decay increases with system scale, consistent with prior findings \cite{dohmatob_strong_2024, herel_collapse_2024, wright_epistemic_2025}. Neural Scaling laws state that test loss depends on model size and dataset size, following a power-law relationship \cite{kaplan_scaling_2020, hoffmann_training_2022}. 


\subsection*{Diversity and EDQ}
The EDQ hypothesis helps explain the impact of ecosystem diversity on model performance. First, in our segmentation approach, sample size per model decreases as diversity increases; thus, once models operate in a low-EDQ regime, greater diversity becomes beneficial. This relationship is consistent with prior work \cite{dohmatob_strong_2024} and reflects a trade-off in long-term considerations of AI ecosystems between large-scale under high-EDQ and small-scale under low-EDQ.
Second, the distribution of an artificial dataset closely resembles the distribution of the model from which it is sampled. If a model had zero sampling error, these distributions would be identical. However, under our hypothesis, data whose distribution closely mirrors the model itself has low EDQ. In contrast, in a diverse ecosystem, different models represent different components of the true distribution. As a result, the  EDQ of redistributed outputs can be substantially higher than in a setting where models only see their own output: data that is not useful for the model that generated it may still be useful for another model within the same ecosystem (see Illustration~\ref{fig:effectivedataquality_EDQ} in the Appendix). This introduces a second trade-off for maximizing EDQ: between larger datasets per model when the data are fresh to the model ($t=0$) and higher EDQ through diversity in later iterations.

We now use the proposed theory to interpret performance dynamics across different stages of self-training iterations.

\textbf{Iteration step $t=0$}
At $t=0$, all models are trained on subsets of real data and therefore operate in a high-EDQ regime. EDQ increases with sample size per model $n = N/M$, which decreases as the number of models $M$ increases. Because default scaling laws apply in this regime, performance decreases with diversity and improves with system scale (see Section~\ref{sec:result_scaling}).

\textbf{Later iterations}
In later iterations, more diverse ecosystems can exhibit lower performance loss (at $t=1$) and/or greater performance gains because (i) the exchange of artificial data among models increases EDQ, and (ii) once models enter the low-EDQ regime, larger per-model sample sizes $n$ become harmful.
This explains the sharp decay observed for $D=1$ in Figure~\ref{fig:both_evolution}, which is further amplified by larger models (Figure~\ref{fig:opt_slice_scaling}). Performance trajectories after $t=0$ illustrate that models in diverse ecosystems receive data of higher EDQ than models in homogeneous ecosystems, ultimately yielding superior long-term performance, despite lower model performance at the beginning. 

Post hoc sensitivity analyses of individual factors that may influence the performance dynamics observed in the evolution of AI ecosystems support our EDQ hypothesis (see Appendix~\ref{sec:sensitivity_analysis}). The results indicate that neither (model-independent) data quality, training data size, the number of models per ecosystem, nor overfitting can fully account for the observed performance decay or the differences across ecosystems.



\section{Discussion}









In our study of AI ecosystem evolution, we find that the optimal level of diversity increases monotonically with the number of self-training iterations (see Figure~\ref{fig:opt_diversity_iterations}).

This offers a new perspective on generative model performance. Training a single model on as much data as possible is short-sighted: over longer horizons, community models trained on segments of the available data outperform a single model with access to the full dataset. The more often artificial data is recycled for training, the greater the degree of segmentation needed to attain the optimal diversity level. Put simply, the benefits of diversity grow with the number of iterations. 

This observed effect is robust across model families, model sizes, dataset scales, and experimental variations involving real data and temperature-based sampling. Moreover, we find that the benefits of diversity also grow as both models and datasets scale, consistent with empirical findings on the relationship between diversity and size in complex systems \cite{yang_scaling_2026}. Taken together, our results suggest that differences between models are a key mechanism for mitigating collapse even under more realistic conditions where models are larger, trained on far more data, can be optimized for diverse outputs, and where human-generated data accumulates alongside model-generated data. Conversely, the results indicate that diversity must increase as the model size and data set grow. \\ 

Before turning to sociotechnical implications for AI evaluation and governance, we briefly restate the technical aspects from Section~\ref{sec:theory}. 
To interpret our empirical results, we introduce \textit{effective data quality} (EDQ), which treats the informativeness of data as a function of both the model distribution and the true distribution\footnote{Importantly, even if the EDQ hypothesis were falsified, the empirically observed diversity optimum would remain.}. When EDQ is sufficiently high, additional training improves performance; when EDQ is low, further training is harmful and performance declines. This framing has two broad implications for recursively trained models and machine learning more generally. First, gains and losses depend on EDQ regardless of whether the data is human- or model-generated \cite{wang_theoretical_2025, feng_beyond_2024, antonello_selecting_2021}. Second, in low-EDQ regimes, standard scaling laws invert: performance degradation increases with scale, consistent with prior findings \cite{dohmatob_strong_2024, herel_collapse_2024, wright_epistemic_2025}. Through the lens of EDQ, diversity helps prevent collapse for two reasons: (i) once the low-EDQ regime is reached, reducing data size per model via segmentation becomes beneficial; and (ii) distributing model-generated data across diverse models increases the model-dependent informativeness of the data each model receives. By contrast, increasing data per model (and reducing diversity) is only beneficial early on, when models are first exposed to new data and EDQ is dominated by sample size. 
The advantage of monoculture observed short-term can even disappear entirely under three modeling variations: introducing diversity by combining model families rather than segmenting data, instantiating diversity via dataset expansion rather than dataset segmentation, and evaluating community models on community-specific test sets rather than a universal benchmark. These findings, reported in the Appendix~\ref{sec:additional_variations}, suggest that various forms of diversification across models may contribute to collapse mitigation. \\

How should these results inform AI goals and governance? Most directly, the systemic risks of AI monoculture challenge the premise of a single universally optimal model, aligning with prior calls for diversity motivated by justice, fairness, and safety \cite{jain_algorithmic_2024, blili-hamelin2025position, gebru_tescreal_2024, lu_model_2024, fisher_political_2025, lazar_ai_2023}. Importantly, promoting diversity in AI does not imply endorsing inferior models; rather, it acknowledges that there are multiple legitimate conceptions of what constitutes beneficial AI.

Incentivizing diversity to mitigate systemic risks is a common approach across many domains. For example, researchers have proposed increasing gender diversity to improve scientific performance \cite{nielsen_gender_2017}, protecting biodiversity to maintain ecological stability \cite{lindenmayer_hidden_2018, otero_biodiversity_2020}, and preserving diversity among financial institutions to reduce the risk of systemic collapse \cite{haldane_systemic_2011, beale_individual_2011, butzbach_systemic_2016}. The financial sector provides useful examples of how such ideas translate to practice, including the U.S. preservation of Minority Depository Institutions \cite{noauthor_statement_2021} and federal tax exemptions for credit unions \cite{noauthor_26_nodate}. Similarly, the EU Treaty explicitly protects the coexistence of banks with different ownership structures and objectives \cite{erkki_liikanen_2012, ferri_how_2018}.

Similar to our proposal, the underlying concern is that correlated behavior arising from agent concentration and/or interdependence increases the risk of systemic failure. 
The AI community has likewise begun discussing policy approaches for promoting diversity across models \cite{fisher_political_2025, jain_algorithmic_2024, dai_embracing_2026}. 

We envision three stages of policies to mitigate the risks of model and knowledge collapse. First, and the most feasible, is to increase transparency regarding system objectives, outputs, and training procedures and data, building on existing approaches such as data statements \cite{bender_data_2018}, datasheets \cite{gebru_datasheets_2021}, and model cards \cite{mitchell_model_2019}. 
Second is to monitor similarity across models at the ecosystem level, drawing from recent methods for measuring model and data homogeneity \cite{xu_echoes_2025, zhang_cultivating_2025}. Such a measure could serve as an early-warning indicator of hidden collapse \cite{lindenmayer_hidden_2018}---effects that benchmarks evaluating short-term performance may miss.
Third is to actively incentivize system differentiation. This could include subsidizing domain-specific AI systems that rely on less-common training data, serve minority communities, and produce outputs that differ from mainstream models. Conversely, regulators could limit excessive concentration and dominance by individual models that amplify data homogenization and systemic risk.
The primary goal of new policies would be to ensure that independent models enrich the training data of future systems. Such diversification can be achieved throughout the production pipeline, including differences in data, model architectures, system prompts, training and post-training, interface design, etc. More generally, differences in objectives naturally diversify outputs across systems.

Given the current trajectory toward AI monoculture \cite{lazar_ai_2023, zhang_cultivating_2025}, a shift toward greater diversity appears necessary to sustain long-term knowledge production.







\bibliographystyle{ACM-Reference-Format}
\bibliography{0bib/ex_zotero}

\newpage
\appendix

\section{Background and related work}

\subsection{Collapse}

Model collapse has been theoretically and/or empirically demonstrated for generative models in various settings \cite{alemohammad_self-consuming_2023, shumailov_ai_2024, herel_collapse_2024, xing_caveats_2025, dohmatob_tale_2024}. The observed effects typically involve some form of performance decay; however, the field has not yet arrived at a general definition of the phenomenon \cite{schaeffer_position_2025}.
Our experimental approach for the single model follows Shumailov et al. \cite{shumailov_ai_2024}, who first formalized model collapse and revealed it in OPT-125m. 
Model collapse arises primarily from three sources of error related to training data quality, the model's capacity to represent and accurately learn a given distribution \cite{shumailov_ai_2024}: statistical approximation error, functional expressivity error, and functional approximation error. 
Prior work show that increasing model and data sizes can amplify collapse \cite{wright_epistemic_2025, herel_collapse_2024, dohmatob_strong_2024, dohmatob_tale_2024, feng_beyond_2024}, thereby contrasting established scaling laws in machine learning. 
Model collapse is not only a threat to generative models accidentally (or inevitably) trained on artificial internet data but also when considering approaches leveraging artificial data for efficiency reasons \cite{bai_constitutional_2022, feng_beyond_2024}. 
Very recent work theorizes that model collapse will result in a collapse of knowledge, defined as the continuous narrowing down of ideas, which will lead to broader epistemic problems for our society \cite{wright_epistemic_2025, peterson_ai_2025, wagner_death_2025}. 

\subsection{Collapse mitigation} %
Because of the epistemic risks of collapse, scholars have developed various strategies to prevent such performance decay, or, put differently, to ensure stability. 
The majority of intervention strategies focus on increasing data quality. This usually involves adding fresh non–model-generated data \cite[e.g.,][]{bertrand_stability_2024, gerstgrasser_is_2024},  filtering or correcting artificial data \cite[e.g.,][]{dohmatob_tale_2024, feng_beyond_2024}, or diversifying training data \cite{zhu_what_2025, wright_epistemic_2025, schaffelder_synthetic_2025}.
While such methods have been shown to mitigate the problem, in most experimental settings they cannot entirely prevent performance decay. In addition, it is often infeasible technically and practically to constantly source with fresh data or reliably separate real from artificial data \cite{xing_caveats_2025, dohmatob_strong_2024}.

\subsection{AI ecosystem}
Our real AI ecosystem is a multi-agent system \cite[e.g.,][]{fu_optimal_2026} comprising a collection of models trained on both their own outputs and those produced by others. Thus, more closely aligned with the work presented here are recent approaches that seek to understand model collapse at the ecosystem level---for instance, by simulating interactions among AI models via retrieval-augmented generation (RAG) \cite{wang_llm_2025}, or by sourcing artificial data from multiple language models for training and evaluating a single model \cite{schaffelder_synthetic_2025}. In contrast, our approach focuses on diversity \textit{across} models as the independent variable, involving real (non-simulated) re-training of language models on their collective output.


 \subsection{Diversity}
 Epistemic diversity in humans refers to different ways of knowing. It broadly describes the phenomenon that individuals with different background assumptions, shaped by diverse lived experiences, values, and beliefs, can arrive at different interpretations when presented with the same information or data \cite{douglas_science_2009, longino1990science, solomon_norms_2006}. 

 AI models generate information according to learned probability distributions, which means each model reflects specific assumptions about what counts as relevant, likely, or meaningful knowledge; in other words, a particular epistemology. Epistemic diversity in an AI ecosystem therefore captures the extent to which multiple models, shaped by distinct data sources, architectures, or objectives, may yield divergent outputs when presented with the same input. 
In recent AI research, epistemic diversity within and across AI models has attracted increased interest. For example, scholars have shown that diverse models collectively make better decisions \cite{kleinberg_algorithmic_2021}, reduce the risk of failures \cite{lu_model_2024}, and enable fairer representations of diverse societies \cite{fisher_political_2025, gebru_tescreal_2024, jain_algorithmic_2024, blili-hamelin2025position}. 
 
Originating in ecology, a common method to measure such diversity is the Hill–Shannon Diversity (HSD) \cite{roswell_conceptual_2021, jost_entropy_2006, hill_diversity_1973, wright_epistemic_2025}, which reflects the effective number of distinct, equally common elements (e.g., models, species, or categories) in a population.

\section{Illustration of model collapse. }
\begin{figure}[t]
    \centering
    \includegraphics[width=0.75\linewidth]{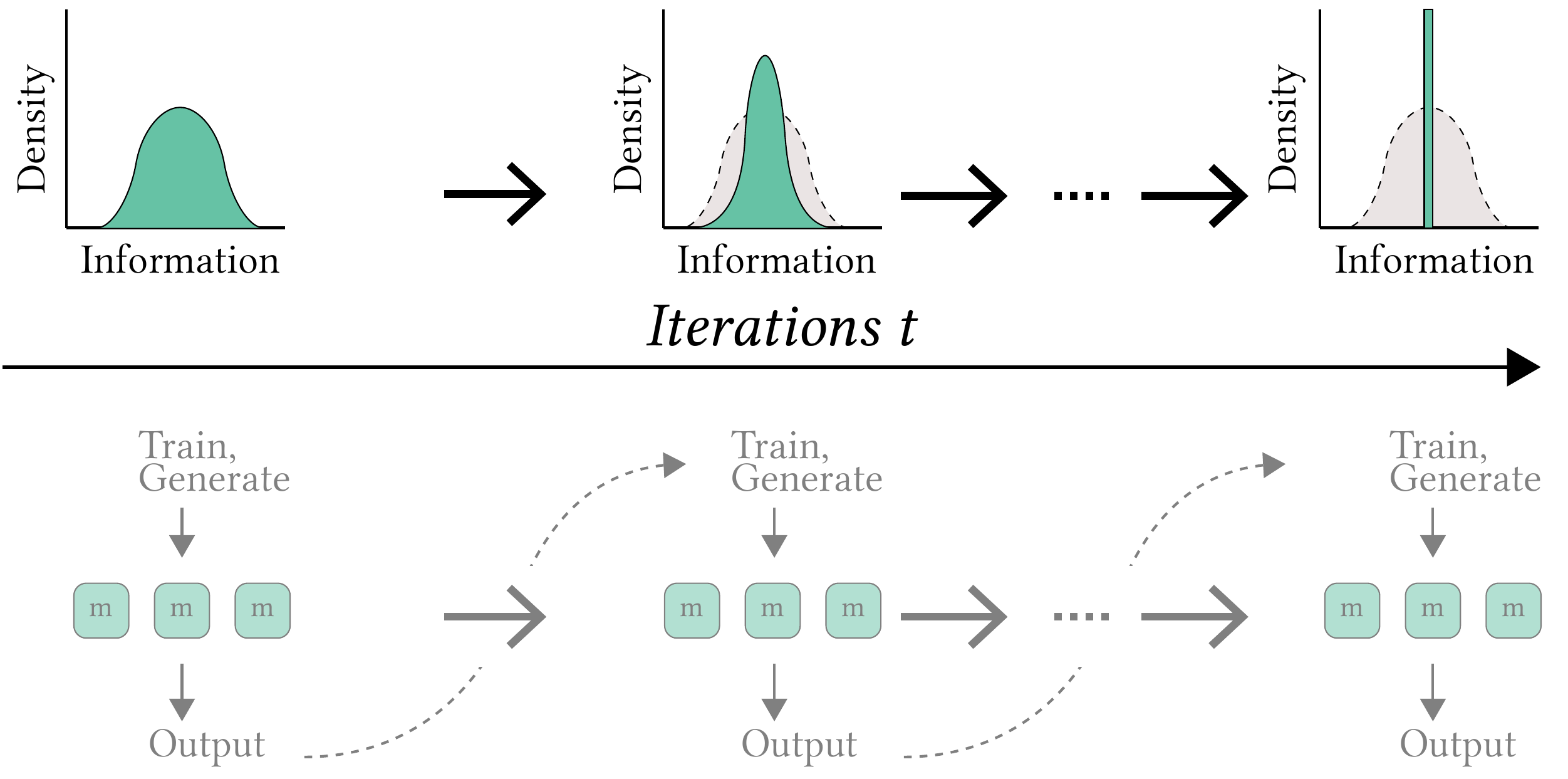}
    \caption{\textbf{Illustration of model collapse in AI ecosystems.} 
    Because of a model's (m) inherent biases (in learning and resampling) and finite data, it cannot perfectly regenerate the information space (e.g., concepts, claims, facts) it was trained on. Some information becomes distorted, and some is lost entirely due to low sampling probabilities. When this training–generation cycle is repeated, the model's output becomes increasingly homogeneous and degrades over iterations $t$. The probability distribution of the information space expressed in the output (and represented by the model, $p_{\theta}$) gradually narrows with increasing $t$, converging toward delta-like functions.
    Collapse can occur at both the individual model level and the ecosystem level, meaning that even a collection of models trained on their collective output can eventually degenerate (here illustrated with three models).
    In society, we expect to advance knowledge. Without human intervention, the best we can hope for from AI models, though, is that they preserve the original distribution of information. Thus, for the purposes of this paper, we evaluate ecosystems with varying levels of diversity based on model performance on a fixed ground-truth dataset over 10 self-training iterations.
    }
    \label{fig:collapse}
\end{figure}

Figure~\ref{fig:collapse} illustrates model collapse in AI ecosystems.
\section{Model progress including performance on base models}
\begin{figure}[htbp]
    \centering
    \begin{subfigure}[b]{0.8\linewidth}
        \centering
        \includegraphics[width=\linewidth]{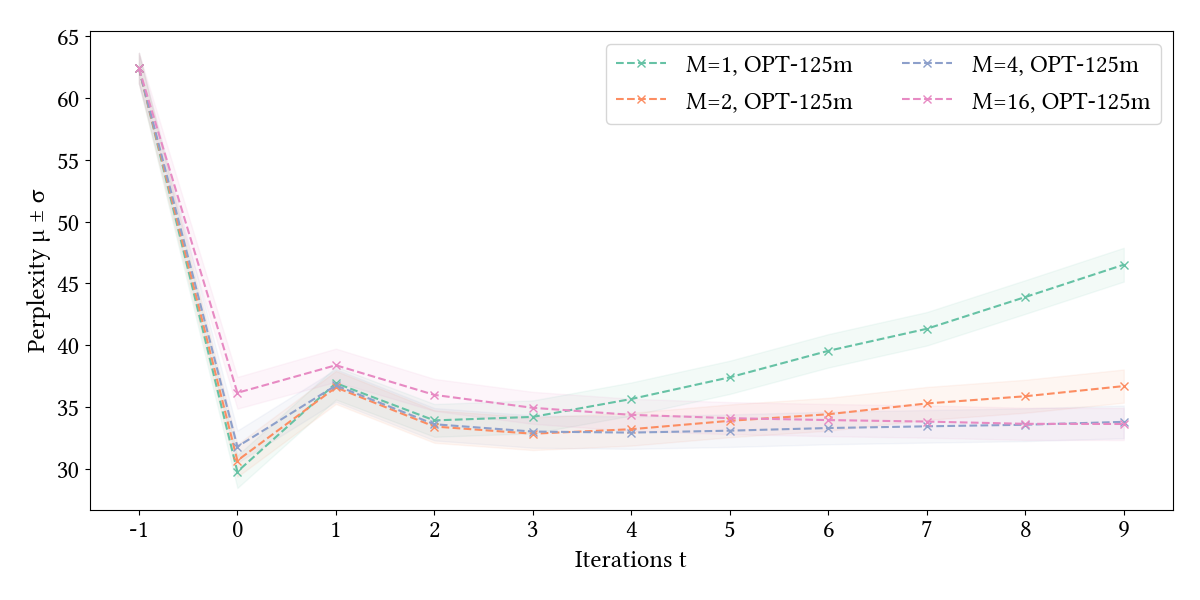}
        \caption{OPT-125m}
    \end{subfigure}
    \\ 
    \begin{subfigure}[b]{0.8\linewidth}
        \centering
        \includegraphics[width=\linewidth]{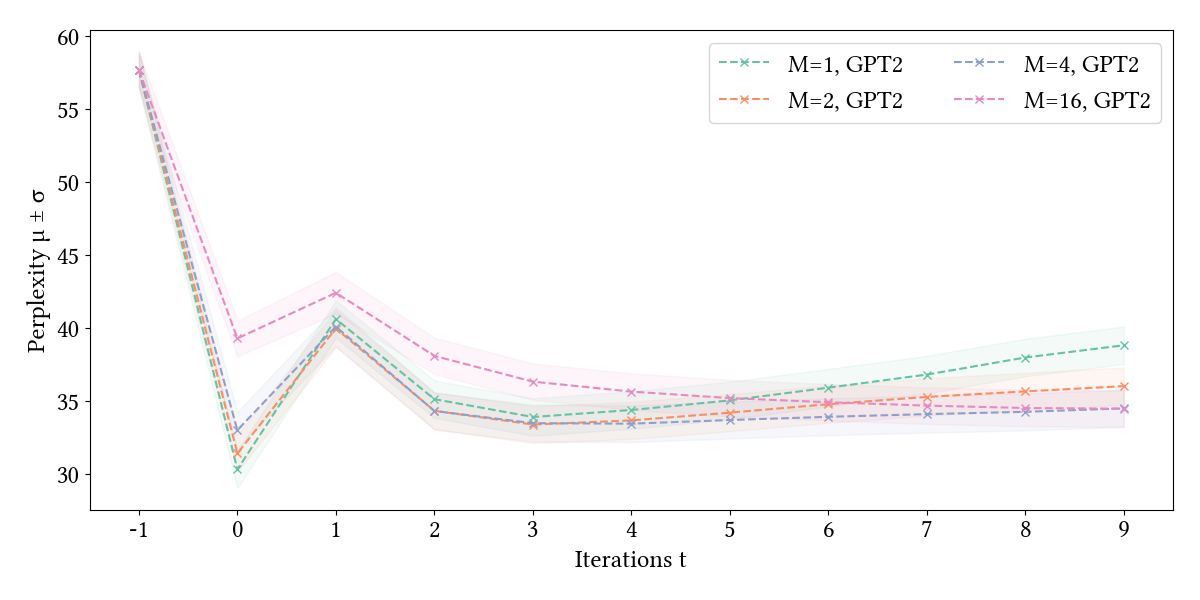}
        \caption{GPT2}
    \end{subfigure}
    \caption{\textbf{Comparison of ecosystems' perplexity.} In this figure, the comparison is made by increasing diversity ($D=M$) across ten iteration steps, including models' perplexity for the off-the-shelf models ($t=-1$). The significant perplexity drop at the first iteration step shows that the fine-tuning mechanism works.  
    }
    \label{fig:both_evolution_start-1}
\end{figure}

Figure~\ref{fig:both_evolution_start-1} shows the perplexity progress of the two model families across ten iteration steps, including the models' perplexity for the base models.

\section{OPT-125m M=1 runs with different random seeds}

\begin{figure}[htbp]
    \centering
        \includegraphics[width=\linewidth]{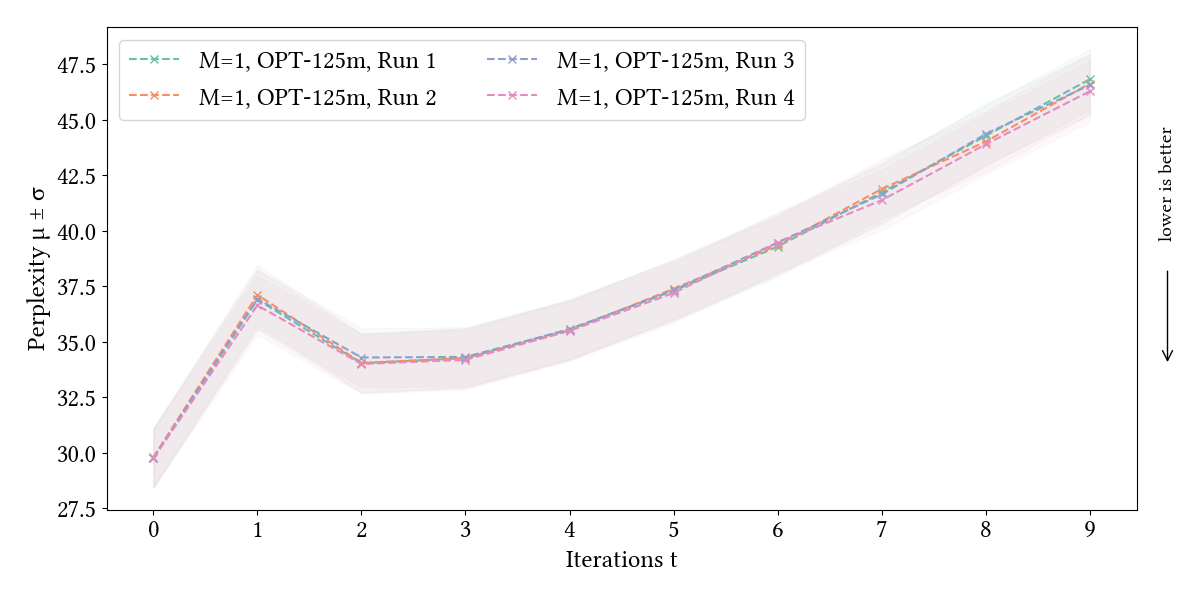}
    \caption{\textbf{The performance trajectory of $M=1$ is largely run-independent.}
    Mean perplexity and standard deviation for ecosystem $M=1$ for each of four independent runs with different random seeds.
    }
    \label{fig:opt_seed_m1}
\end{figure}

Figure~\ref{fig:opt_seed_m1} shows the performance trajectories for $M=1$ across ten iterations for four independent runs. The similarity of these trajectories indicates that the results are largely independent of the random seed. Ecosystems with higher diversity already average performance across two or more models.

\section{OPT-125m performance across 20 iterations}
\begin{figure}[htbp]
    \centering
    \begin{subfigure}[b]{0.8\linewidth}
        \centering
        \includegraphics[width=\linewidth]{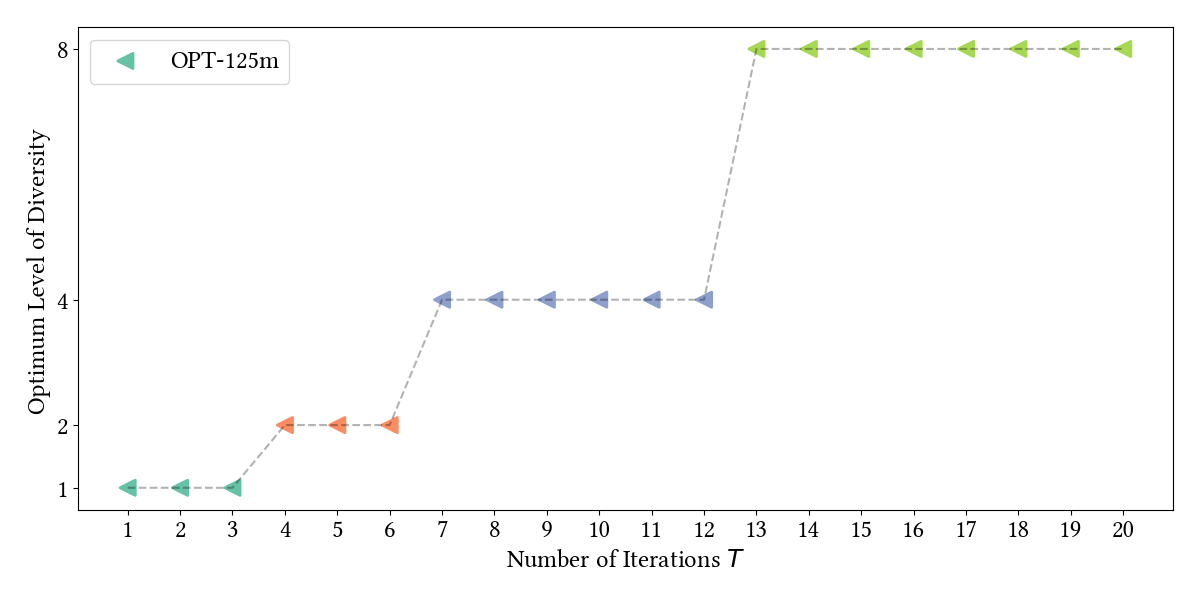}
        \caption{Optimal level of diversity.}
    \end{subfigure}
    \\ 
    \begin{subfigure}[b]{0.8\linewidth}
        \centering
        \includegraphics[width=\linewidth]{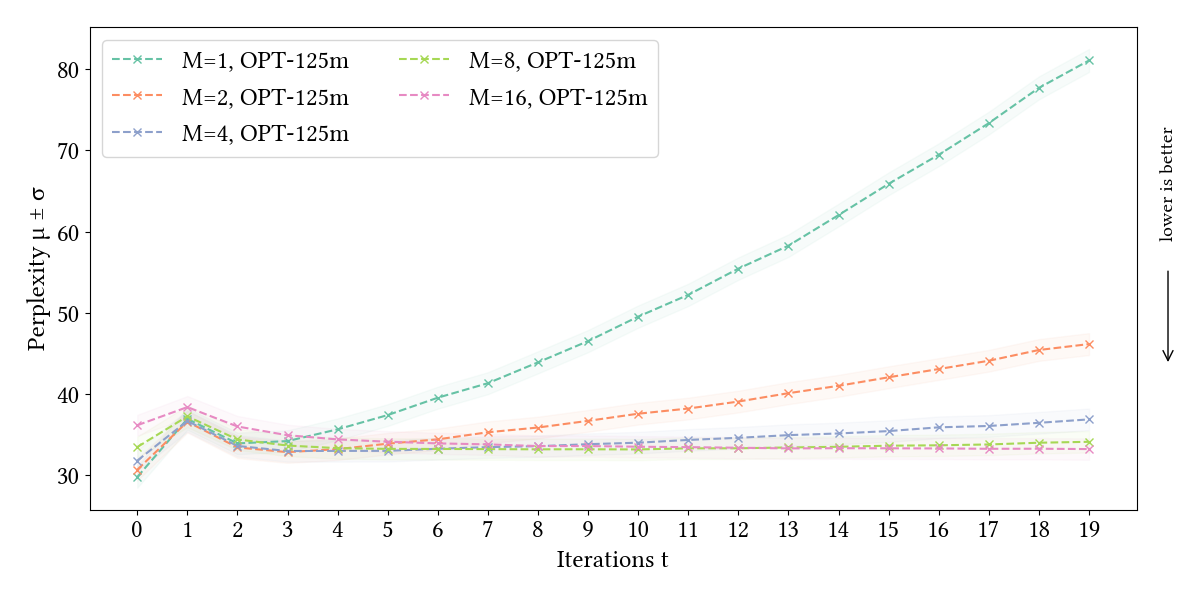}
        \caption{Performance progress.}
    \end{subfigure}
    \caption{\textbf{The optimal level of ecosystem diversity continues to increase beyond 10 self-training iterations ($T$).} 
    (a) Optimal diversity level and 
    the corresponding performance trajectories for OPT-125M. (a) and (b) extend the iteration range of Figures~\ref{fig:opt_diversity_iterations} and \ref{fig:both_evolution}. 
    Colored symbols correspond to the diversity levels shown in Figure~\ref{fig:approach} with an additional level of $D=8$ in light green.
    }
    \label{fig:opt_20}
\end{figure}

Figure~\ref{fig:opt_20} shows the optimal diversity level for OPT-125m across up to 20 self-training iterations ($T$), showing that the optimal diversity level continues to increase beyond 10 iterations.

\section{Illustrations of probability distributions and EDQ}
 
\begin{figure}[ht]
    \centering
    \includegraphics[width=0.75\linewidth]{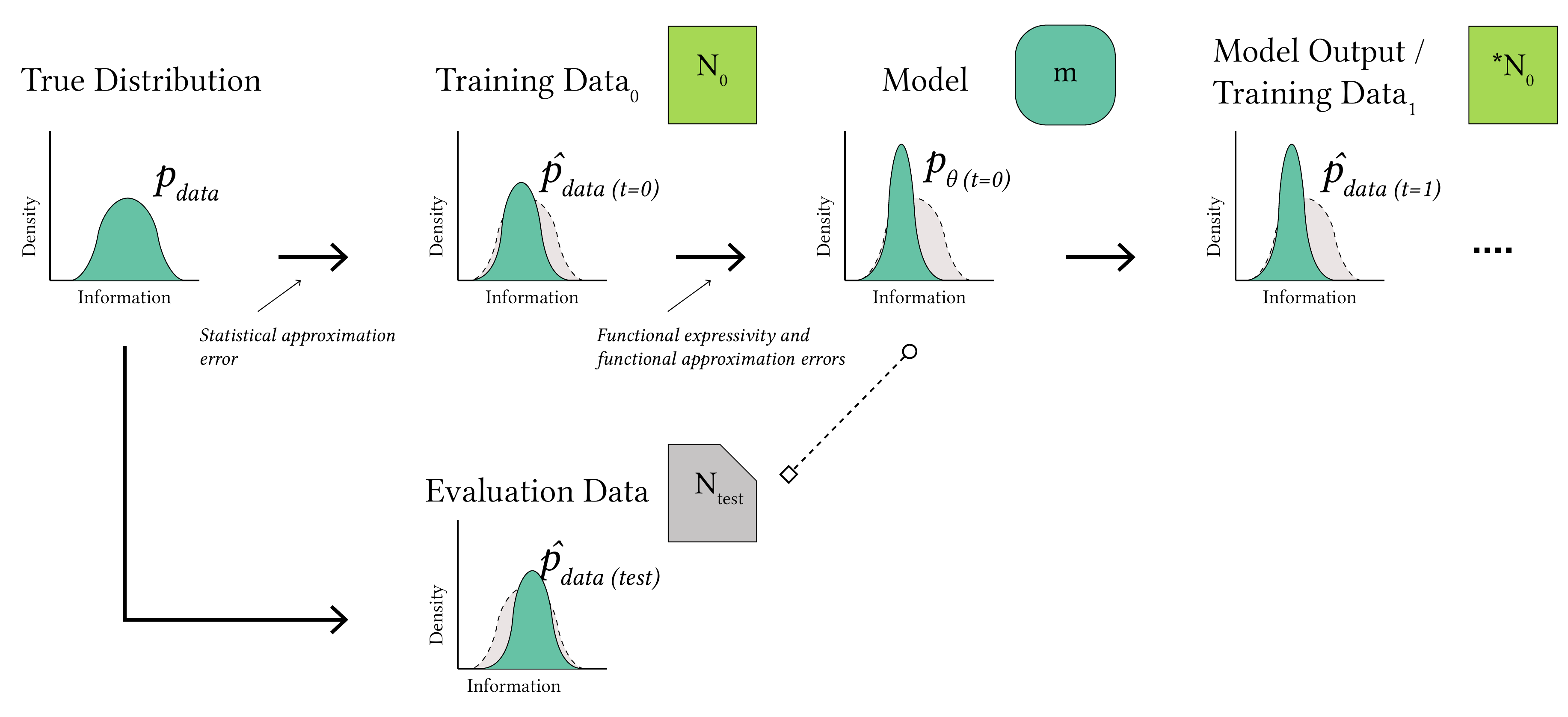}
    \caption{\textbf{Illustration of probability distributions in self-training loops.} Because the training data is finite, the empirical distribution, $\hat{p}_{data (t=0)}$, diverges from the true distribution, ${p}_{data}$. The model's inherent biases (functional expressivity and functional approximation errors) further distort the distribution it represents, which is $p_{\theta (t=0)}$, and the distribution of the data it generates, which is $\hat{p}_{data (t=1)}$.
    }
    \label{fig:distributions}
\end{figure}

\begin{figure}[ht]
    \centering
    \includegraphics[width=0.75\linewidth]{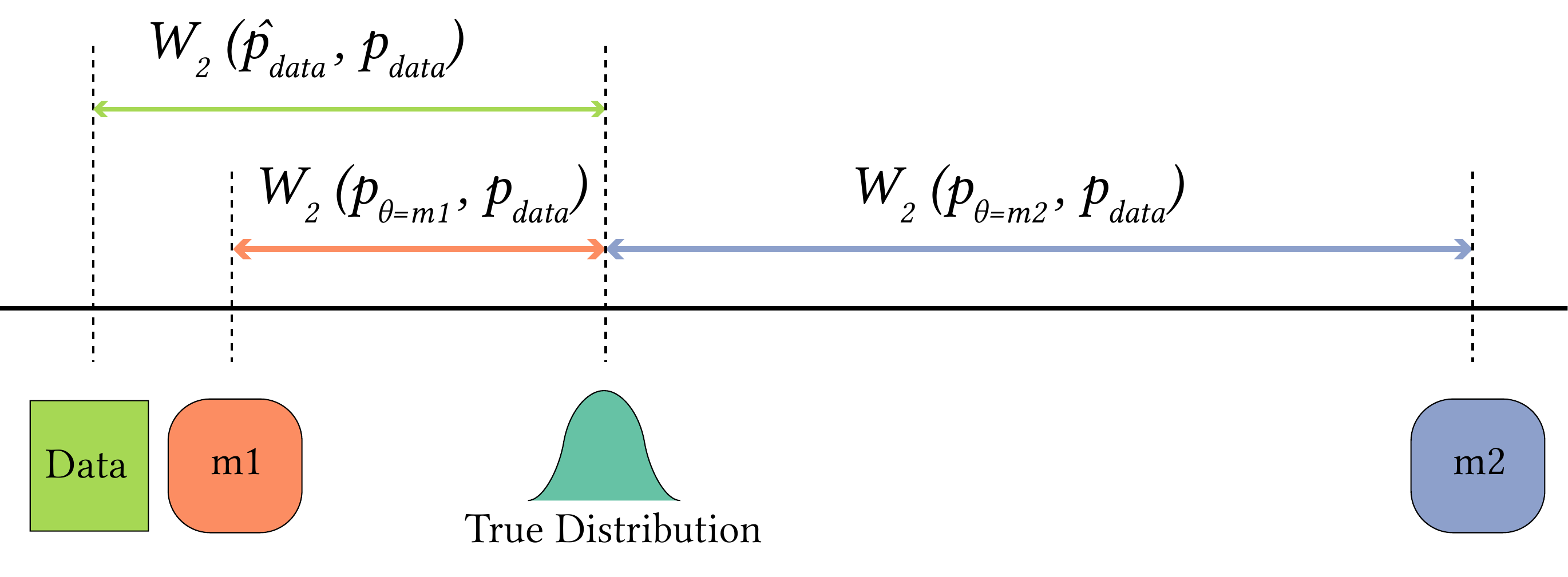}
    \caption{\textbf{Illustration of model-dependent EDQ.} For the illustrated distribution discrepancies, the training data (``DATA'') is beneficial for ``m2'' (high EDQ) but harmful for ``m1'' (low EDQ). EDQ is defined as
    $\delta_{EDQ}(\hat{p}{data}, p{\theta}) := \mathbb{W}2(\hat{p}{data(t)}, p_{data}) - \mathbb{W}2(p{\theta(t)}, p_{data}),$
    where lower values of $\delta_{EDQ}$ indicate higher EDQ. It is important to note that the introduced scalar metric serves as a heuristic to explain model-dependent data quality; therefore, we ignore additional factors shaping model weights.
    }
    \label{fig:effectivedataquality_EDQ}
\end{figure}

Figure~\ref{fig:distributions} illustrates the probability distributions in self-training loops, and Figure~\ref{fig:effectivedataquality_EDQ} illustrates the model-dependent EDQ.

\section{Perplexity distributions for both model families}
\begin{figure}[htbp]
    \centering
    \begin{subfigure}[b]{0.8\linewidth}
        \centering
        \includegraphics[width=\linewidth]{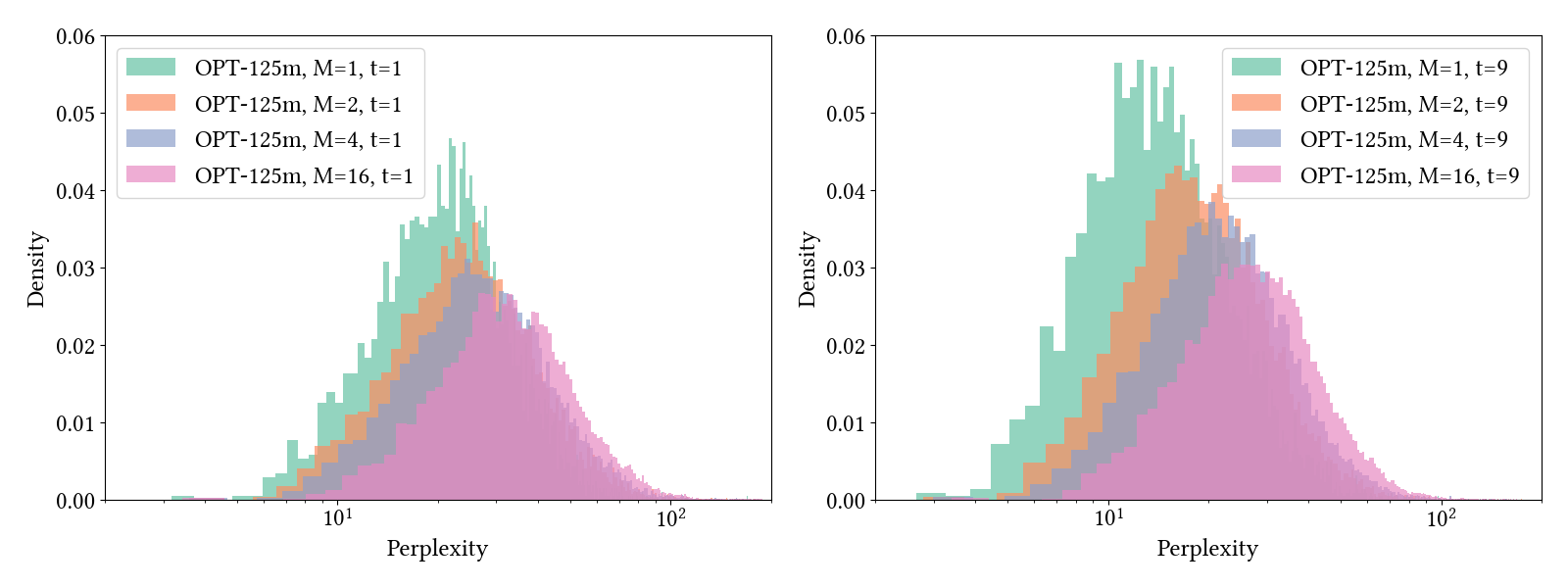}
        \caption{OPT-125m}
        \label{fig:opt_dist}
    \end{subfigure}
    \\ 
    \begin{subfigure}[b]{0.8\linewidth}
        \centering
        \includegraphics[width=\linewidth]{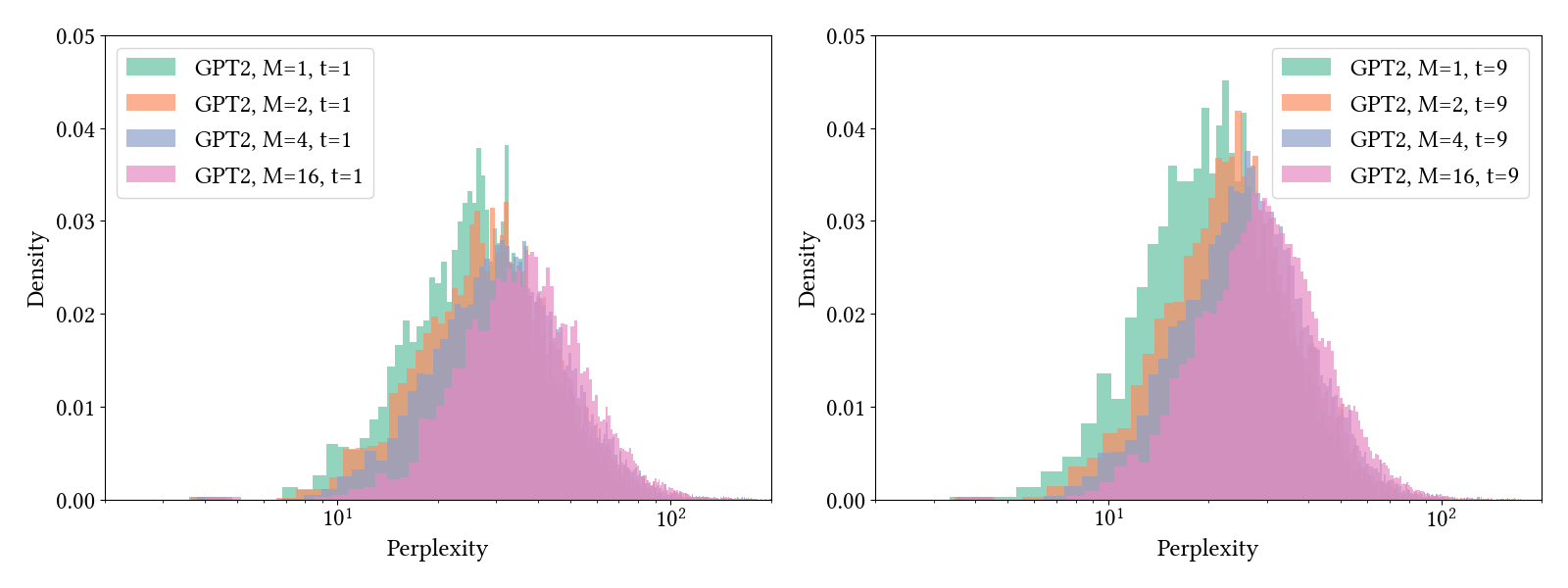}
        \caption{GPT2}
        \label{fig:gpt2_dist}
    \end{subfigure}
    \caption{\textbf{Perplexity distributions widen with increasing diversity.} Mixed perplexity distribution on the original Wikitext2 train data points, evaluated by all models in a given ecosystem with diversity ($D$ =$M$ = 1, 2, 4, 16) at iteration steps $t$ = 0 (left) and $t$ = 9 (right). As number of models increases, the distributions become wider, indicating greater diversity. Simultaneously, their means shift toward higher perplexity, suggesting greater divergence from the distribution in the original training set. Across all ecosystems, the distributions narrow from the beginning to the end of the model iterations. 
    }
    \label{fig:both_dist_v0}
\end{figure}

Figure~\ref{fig:both_dist_v0} shows the mixed perplexity distributions of the  OPT-125m and GPT2 ecosystems on the original training dataset, indicating that model collections with greater diversity exhibit both greater variety in the information represented and lower alignment with the training data distribution.

\section{GPT2 V1: Model size and data size}
\label{sec:gpt2_scaling}

\begin{figure}[htbp]
    \centering
    \begin{subfigure}[t]{0.5\textwidth}
        \centering
        \includegraphics[width=\linewidth]{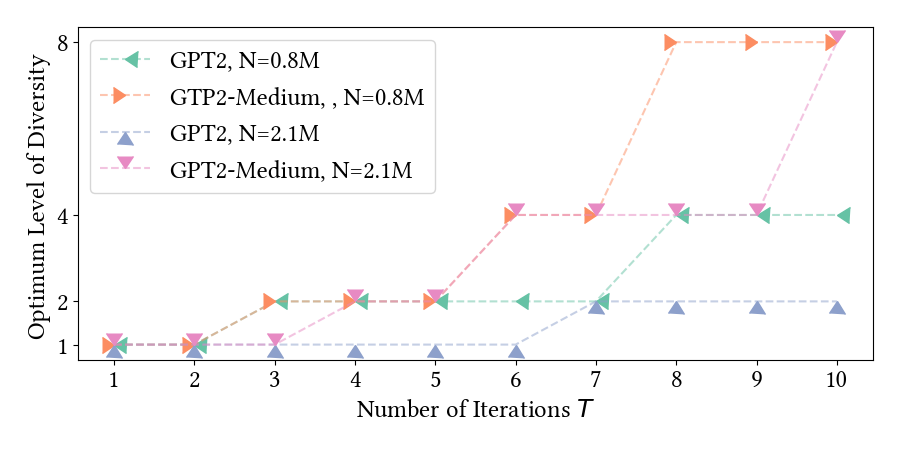}
        \caption{Optimal diversity levels.}
    \end{subfigure}
    ~ 
    \begin{subfigure}[t]{0.5\textwidth}
        \centering
        \includegraphics[width=\linewidth]{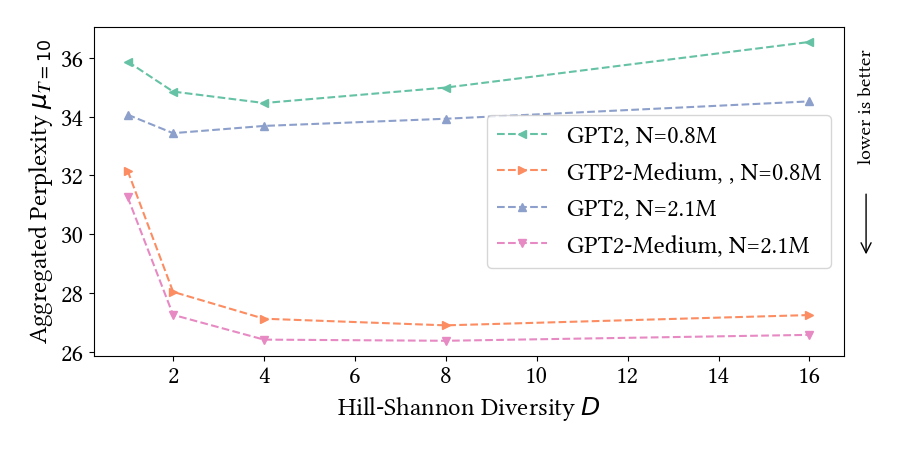}
        \caption{Aggregated perplexities after 10 iterations. 
        }
    \end{subfigure}
    \\ 
    \centering
    \begin{subfigure}[t]{0.5\textwidth}
        \centering
        \includegraphics[width=\linewidth]{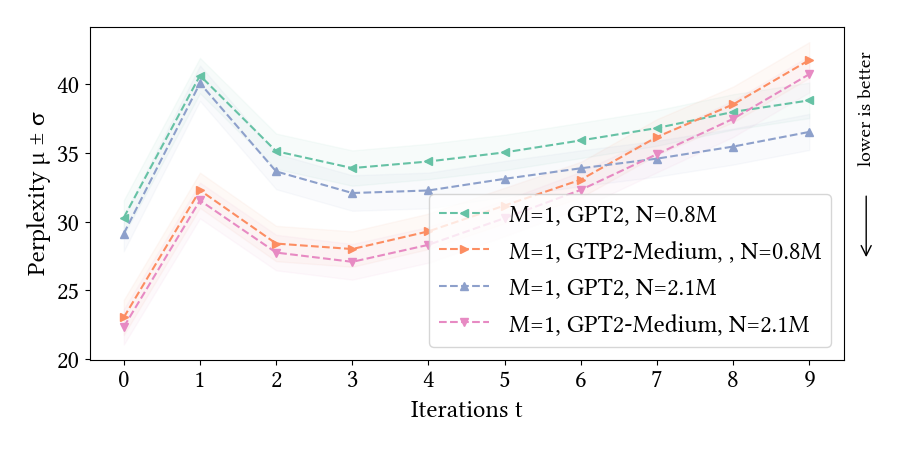}
        \caption{Mean perplexities for M;D=1}
    \end{subfigure}
    ~ 
    \begin{subfigure}[t]{0.5\textwidth}
        \centering
        \includegraphics[width=\linewidth]{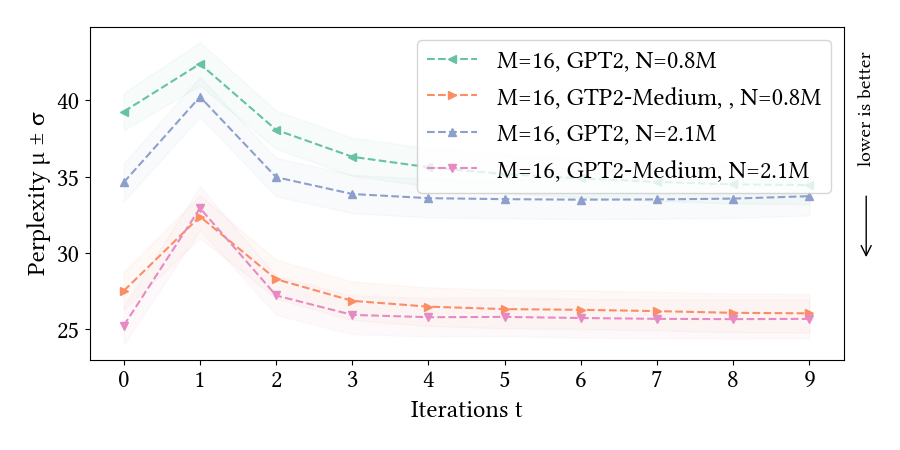}
        \caption{Mean perplexities for M;D=16}
    \end{subfigure}
\caption{
\textbf{The optimal level of diversity for GPT-2 performance increases as both model and dataset size grow.}
Ecosystem diversity across four system scales (two model sizes × two dataset sizes; variation V1). ``GPT-2, $N=0.8$M'' (green $\blacktriangleleft$) corresponds to the primary setting, using a 124M-parameter model and a total training dataset of $0.8$M tokens. ``GPT-2 Medium'' corresponds to the 355M-parameter model in the GPT-2 family.
\textbf{Top:} (a) Optimal diversity level vs.\ number of iterations $T$ and (b) aggregated perplexity $\mu_{T=10}$ for five diversity levels ($M;D=1,2,4,8,16$). Larger models (orange and pink vs.\ green and blue) show a steeper increase in optimal diversity with $T$, shifting the optimum from $D=4$ to $D=16$ after 10 iterations. At a fixed model size, increasing dataset size (pink $\blacktriangledown$ and blue $\blacktriangle$ vs.\ green and orange) decreases the optimal level.
\textbf{Bottom:} Perplexity trajectories across ten iterations for two diversity levels (c: $D=1$, d: $D=16$). For a fixed diversity level, increasing model or dataset size reduces perplexity at $t=0$.
(c) In the most homogeneous ecosystem ($D=1$), larger models show strong perplexity increases in later iterations, indicating stronger collapse, while solely increasing dataset size slightly decreases perplexity.
(d) In highly diverse ecosystems ($D=16$), increasing model and/or dataset size decreases perplexity. 
The shaded region represents the combined standard deviation across models and data points.}
\label{fig:gpt2_slice_scaling}
\end{figure}

Figure~\ref{fig:gpt2_slice_scaling} shows GPT2 performance results for increased model size (GPT2-Medium) and larger data size ($2.1$M tokens).

\section{Additional experimental variations of the primary setting}
\label{sec:additional_variations}

\begin{figure}[ht]
    \centering
    \includegraphics[width=0.75\linewidth]{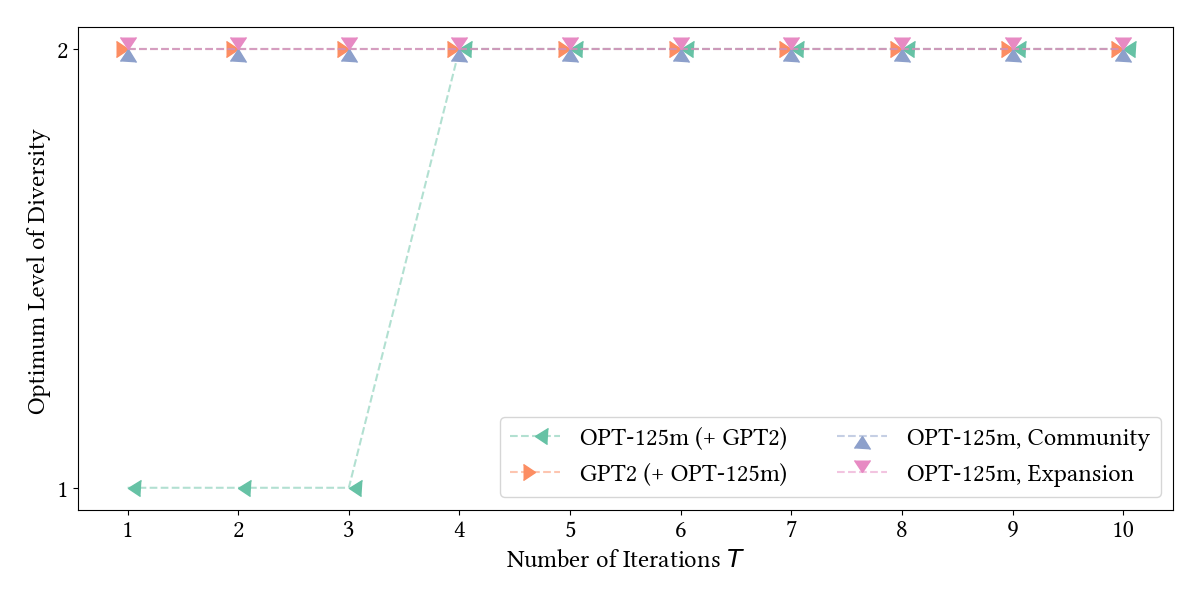}
    \caption{\textbf{The advantage of homogeneous ecosystems observed short-term can even disappear entirely under three modeling variations:} introducing diversity by combining model families rather than segmenting data (``OPT-125m(+GPT2)'' and ``GPT2(+OPT-125m)'') instantiating diversity via dataset expansion rather than dataset segmentation (``OPT-125m, Expansion''), and evaluating community models on community-specific test sets rather than a universal benchmark (``OPT-125m, Community''). These variations involved only two diversity levels ($D=1,2$).
    }
    \label{fig:opt_optimumdiversity_variations2}
\end{figure}

\begin{figure}[htbp]
    \centering
    \begin{subfigure}[b]{0.8\linewidth}
        \centering
        \includegraphics[width=\linewidth]{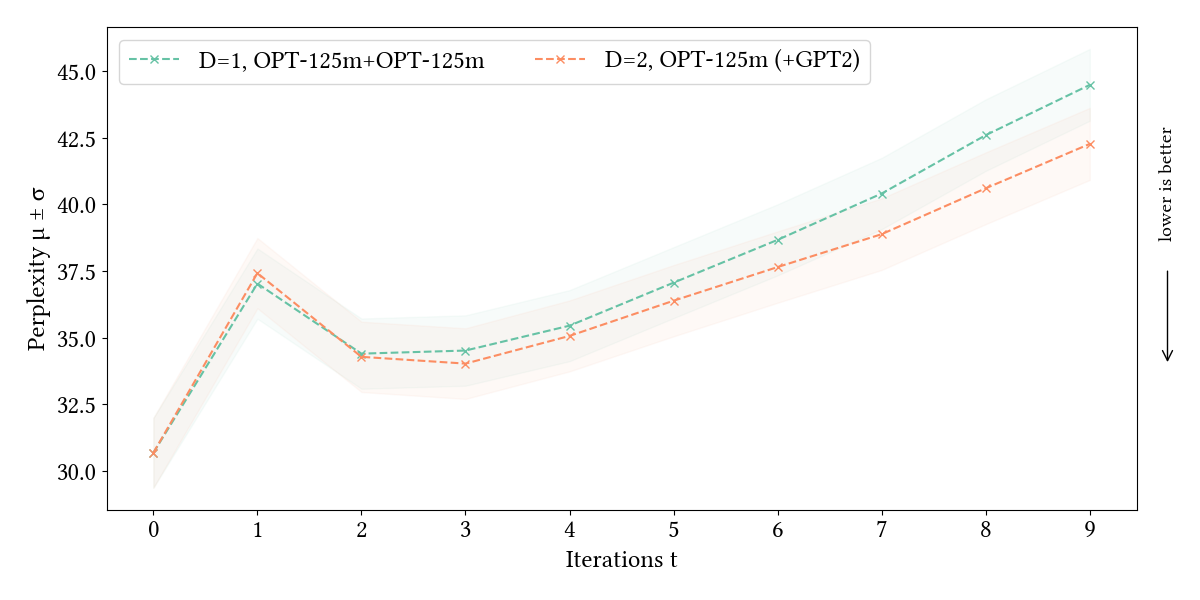}
        \caption{Combination of two OPT-125m models compared with an OPT-125m paired with GPT-2}
        \label{fig:opt_gpt2_family}
    \end{subfigure}
    \\ 
    \begin{subfigure}[b]{0.8\linewidth}
        \centering
        \includegraphics[width=\linewidth]{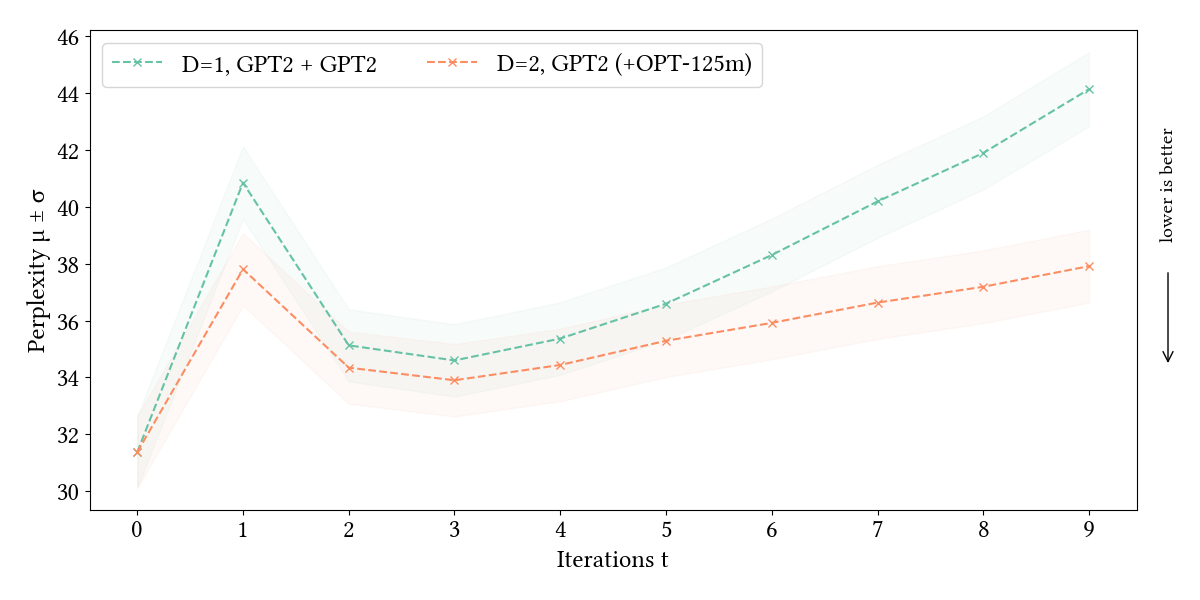}
        \caption{Combination of two GPT-2 models compared with an GPT-2 paired with OPT-125m}
        \label{fig:gpt2_opt_family}
    \end{subfigure}
    \caption{\textbf{Two diverse pre-trained models yield lower perplexity than two identical pre-trained models.}
OPT-125m+OPT-125m and GPT2+GPT2 represent ecosystems composed of two identical models (green). OPT-125m(+GPT2) and GPT2(+OPT-125m) represent ecosystems where each model is paired with the other type (orange).
Unlike the primary segmentation approach, diversity here is instantiated by using two different pre-trained models. All models are trained on the same dataset at iteration step $t=0$ and use the same generation dataset throughout; thus, aside from their pre-training differences, the training procedure is identical across models.
The shaded region represents the combined standard deviation across models and data points.}
    \label{fig:diverse_pretrained}
\end{figure}

\begin{figure}[htbp]
    \centering
    \begin{subfigure}[b]{0.8\linewidth}
        \centering
        \includegraphics[width=\linewidth]{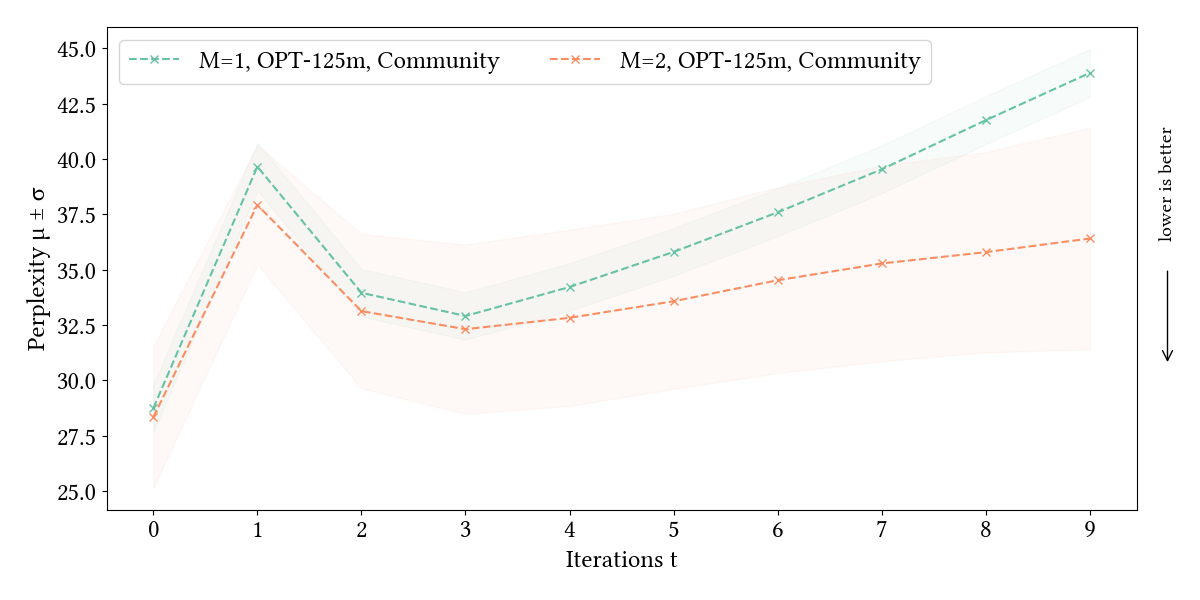}
        \caption{\textbf{From the first iteration step ($t=0$) onward, two community models consistently achieve lower perplexity than a single model trained and evaluated on the combined datasets.}
        $M=1$: a single model trained and evaluated on the combination of Wikitext2 and Penn Treebank (green). $M=2$: two community models, each trained and evaluated on its corresponding dataset (orange).
        To ensure a controlled comparison, the length of the Wikitext2 dataset is reduced to match the length of Penn Treebank. }
        \label{fig:opt_community}
    \end{subfigure}
    \\ 
    \begin{subfigure}[b]{0.8\linewidth}
        \centering
        \includegraphics[width=\linewidth]{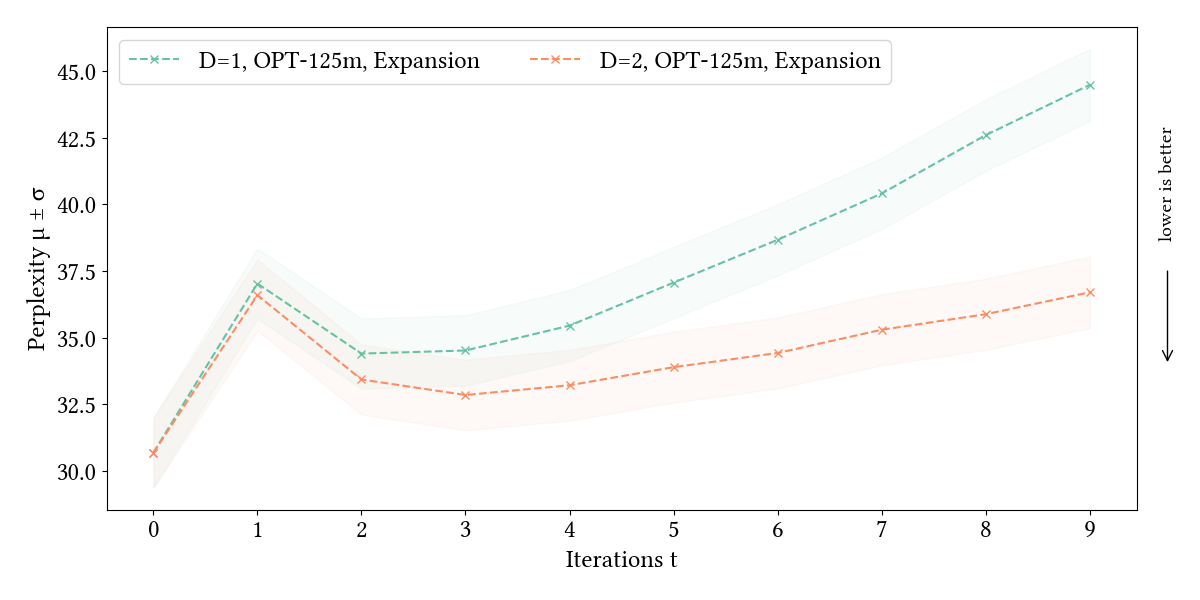}
        \caption{
        \textbf{Two models trained on two different segments of  Wikitext2 consistently achieve lower perplexity than two models both trained on one of the two segments.} $D=1$: Two models trained on identical segment 0 (green). $D=2$: Two models trained on two different segments 0 and 1 (orange), corresponding to the same ecosystem as M=2 in the primary setting. Unlike in the primary segmentation approach, diversity is achieved through expansion of the distribution.}
        \label{fig:opt_expansion}
    \end{subfigure}
    \caption{The shaded region is a combined standard deviation across the models and data points.}
    \label{fig:community_expansion}
\end{figure}

The observed short-term benefit of homogeneous ecosystems is due to the segmentation approach, which reduces the data size per model as diversity increases, and to the use of a universal ground-truth dataset (the Wikitext2 test set in our case). This initial advantage disappears when diversity is instantiated through combining two model families (Figure~\ref{fig:diverse_pretrained}), through data expansion (Figure~\ref{fig:opt_expansion}), or when community-specific test sets are employed. The aggregated results are shown in Figure~\ref{fig:opt_optimumdiversity_variations2}.

\paragraph*{Diversity through combining 2 model families}
Real-world diversity may arise from multiple system-design dimensions, including pre-training data, architecture, alignment, interface design, and system messaging. Figure~\ref{fig:diverse_pretrained} shows that combining two model families (OPT-125m and GPT2) consistently outperforms the combination of two models of same family, suggesting that diversity across design dimensions may further amplify benefits. 

\paragraph*{Diversity via data expansion}
The expansion approach reflects a more realistic landscape of multiple similar models \cite{lazar_ai_2023, zhang_cultivating_2025}. If dominant models are trained on largely overlapping internet-scale data \cite{villalobos_position_2024, mozilla_mozilla_2024}, segmentation through reduction may be the only feasible diversification strategy; otherwise, diversity could be achieved via expansion by incorporating models trained on previously unused data. In such comparisons, diverse ecosystems consistently achieve lower perplexity across iterations (Figure~\ref{fig:opt_expansion}), whereas simply adding more similar models does not prevent collapse (Figure~\ref{fig:opt_control}).

\paragraph*{Community-specific test sets}
Evaluating community models using community-specific datasets rather than universal benchmarks yields a more realistic assessment; under this setting, diverse ecosystems outperform homogeneous ones already at $t=0$ (Figure~\ref{fig:opt_community}). Community- or domain-specific models challenge the prevailing assumption that AI systems should serve everyone equally across all tasks \cite{blili-hamelin2025position, bender_dangers_2021}. Granting affected communities greater agency in AI design \cite{costanza-chock_design_2020} may naturally guide ecosystems toward healthier diversity.

Human language and knowledge evolve alongside AI systems. Following prior work \cite[e.g.,][]{herel_collapse_2024, shumailov_ai_2024, alemohammad_self-consuming_2023}, we use a static dataset for generation and evaluation; in practice, this would resemble evaluating future models on today's data. A more realistic approach would involve evolving generation and evaluation datasets over time.

While further research is needed to identify optimal ecosystem parameters under realistic conditions, our results provide strong evidence that diversity is a key factor in sustaining long-term AI ecosystem performance.

\section{Sensitivity analysis}
\label{sec:sensitivity_analysis}

\begin{figure}[htbp]
    \centering
    \begin{subfigure}[t]{0.5\textwidth}
        \centering
        \includegraphics[width=\linewidth]{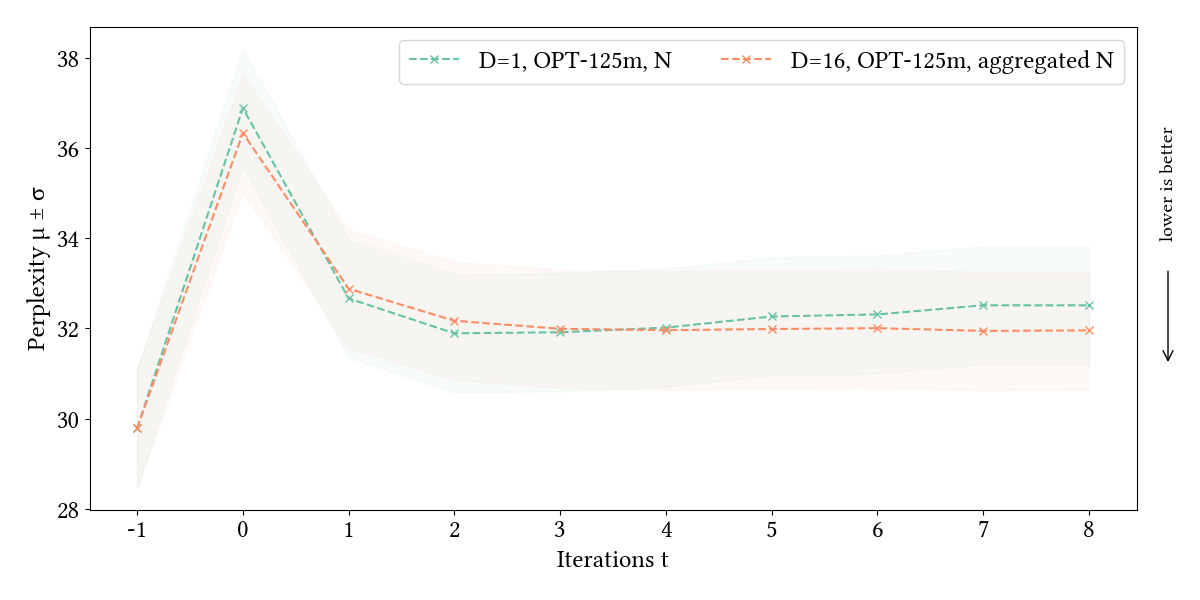}
        \caption{OPT-125m separately trained on generated output}
        \label{fig:opt_dataquality_train}
    \end{subfigure}
    ~ 
    \begin{subfigure}[t]{0.5\textwidth}
        \centering
        \includegraphics[width=\linewidth]{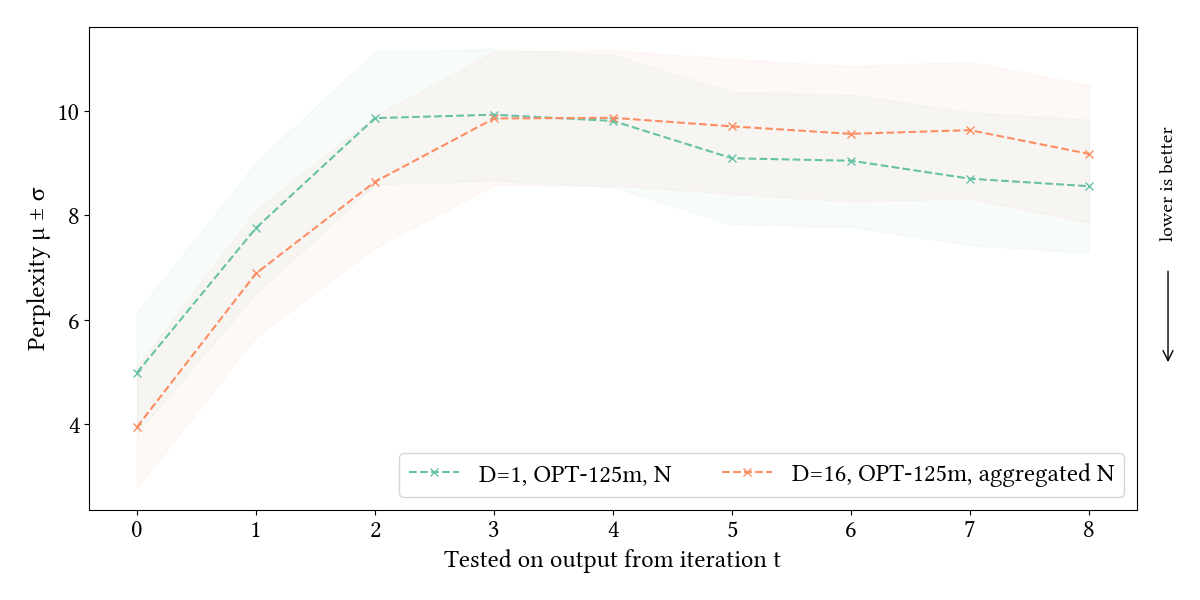}
        \caption{OPT-125m tested on generated output}
        \label{fig:opt_dataquality_test}
    \end{subfigure}
    \caption{\textbf{Data quality alone cannot explain performance differences across ecosystems.}
    The OPT-125m base model was separately trained (a) and tested (b) on generated outputs from the least and most diverse ecosystems ($D=1$ and $D=16$, respectively). For $D=16$, outputs from all 16 models were aggregated. In (a), the off-the-shelf OPT-125m base model was fine-tuned separately on each generated dataset and subsequently evaluated on the fixed Wikitext2 test set. 
    The similar perplexity values obtained for $D=1$ and $D=16$ indicate that model-independent data quality cannot fully account for the performance differences observed in our main experiments. Lower perplexity values for outputs from iterations $t=0$ and $t=1$ in (b) suggest that artificial data generated at early stages of the self-training loop more closely resembles the pretraining distribution of the base OPT-125m model. Fine-tuning on such artificial data shifts the distribution of the fine-tuned model back toward the pretraining distribution and away from the true distribution, which helps explain the performance drop observed with the output from iteration $t=0$ in (a) and consistently across our experiments.}
    \label{fig:control_dataquality}
\end{figure}

\begin{figure}[htbp]
    \centering
    \begin{subfigure}[t]{0.5\textwidth}
        \centering
        \includegraphics[width=\linewidth]{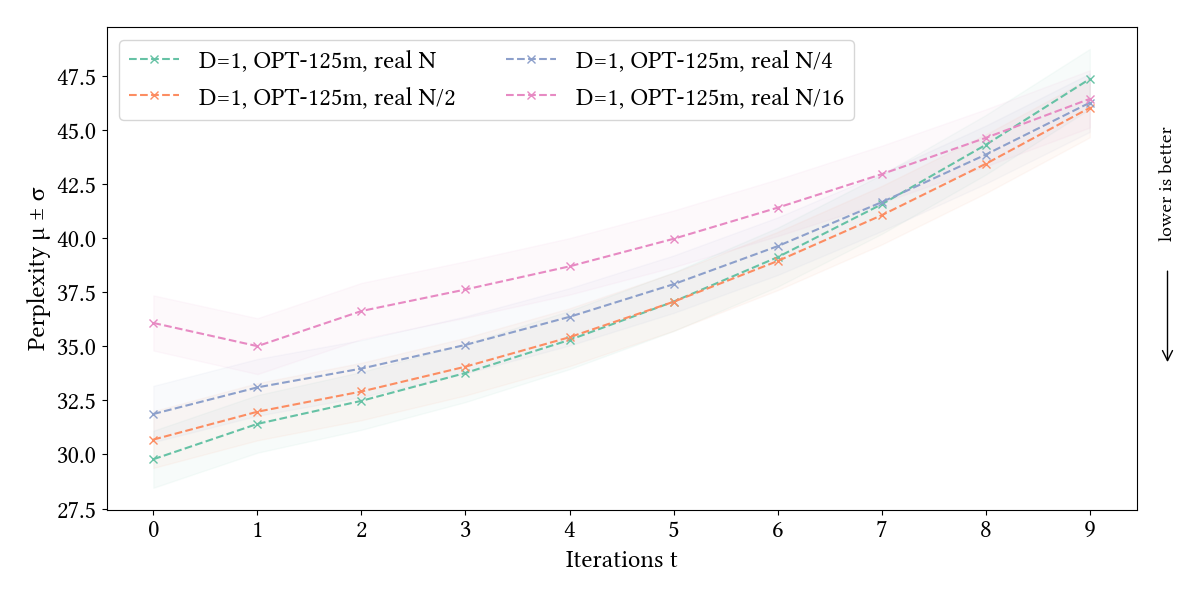}
        \caption{Single OPT-125m re-trained on the real Wikitext2 train set}
        \label{figopt_baselineopt_control}
    \end{subfigure}
    ~ 
    \begin{subfigure}[t]{0.5\textwidth}
        \centering
        \includegraphics[width=\linewidth]{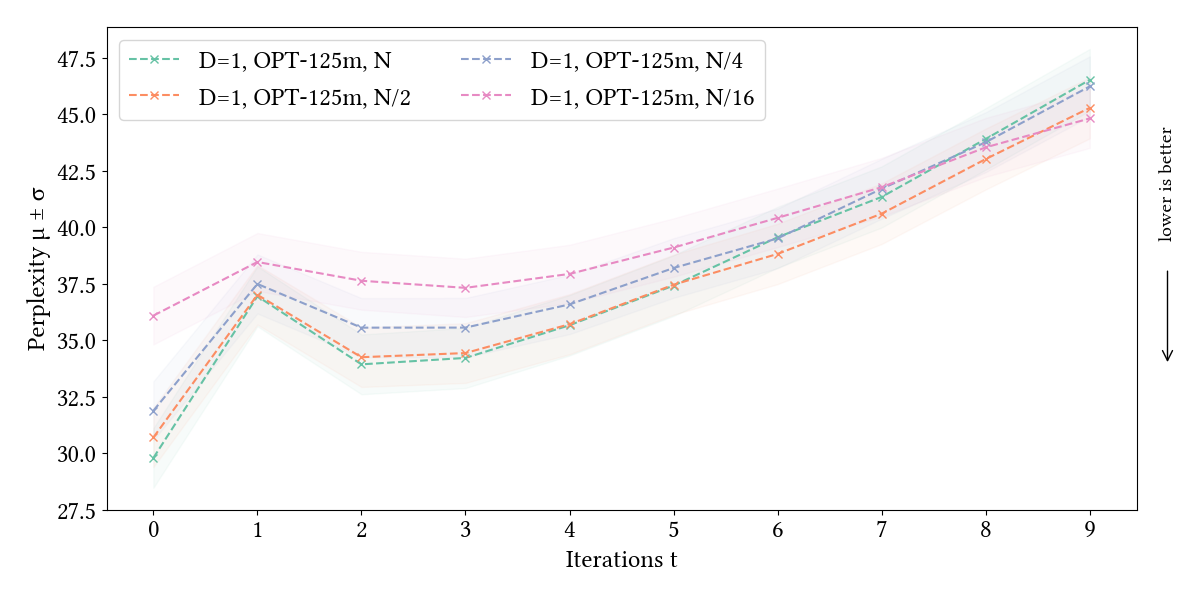}
        \caption{Single OPT-125m self-trained on own output}
        \label{fig:opt_onslice}
    \end{subfigure}
    \caption{\textbf{Training data size alone cannot explain performance differences across ecosystems.} A single OPT-125m model was analyzed in two conditions: (a) re-trained on the real Wikitext2 training set and (b) self-trained on its own output, respectively. The initial training data content and sizes correspond to the primary setting ($N=0.8$M tokens), but only one model was used in each run. The condition ``$D=1$, OPT-125m, $N$'' matches $D=1$ in the primary setting. 
    The similar perplexity values observed across different data sizes indicate that training data size alone cannot account for the performance differences across ecosystems.
        }
    \label{fig:control_datasize}
\end{figure}

\begin{figure}[htbp]
    \centering
    \begin{subfigure}[t]{0.5\textwidth}
        \centering
        \includegraphics[width=\linewidth]{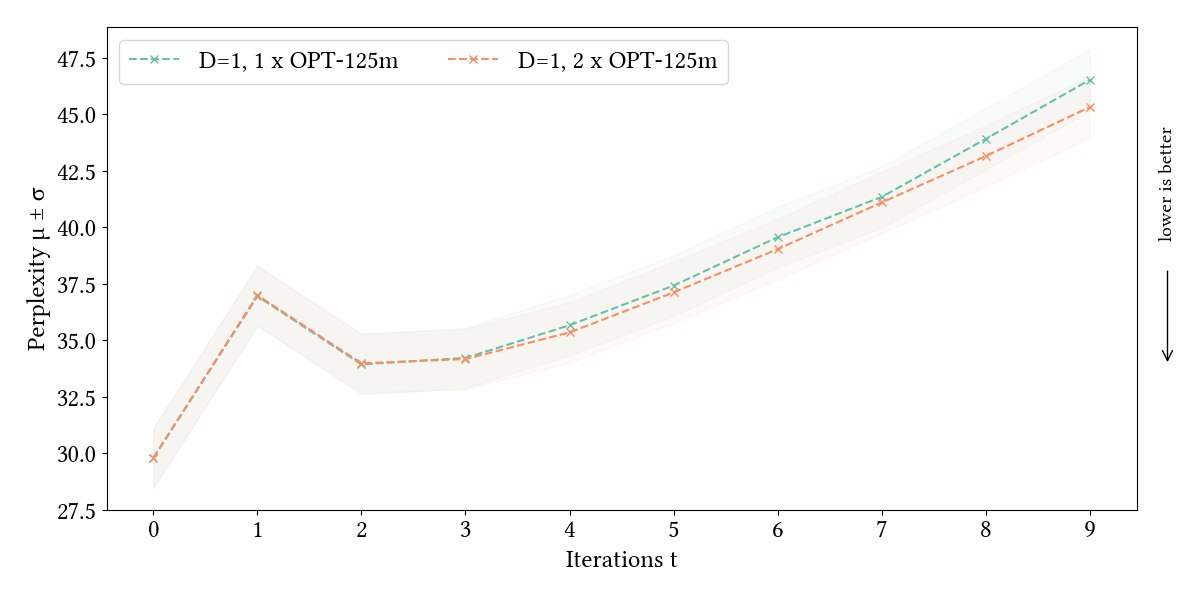}
        \caption{OPT-125m}
        \label{fig:opt_control}
    \end{subfigure}
    ~ 
    \begin{subfigure}[t]{0.5\textwidth}
        \centering
        \includegraphics[width=\linewidth]{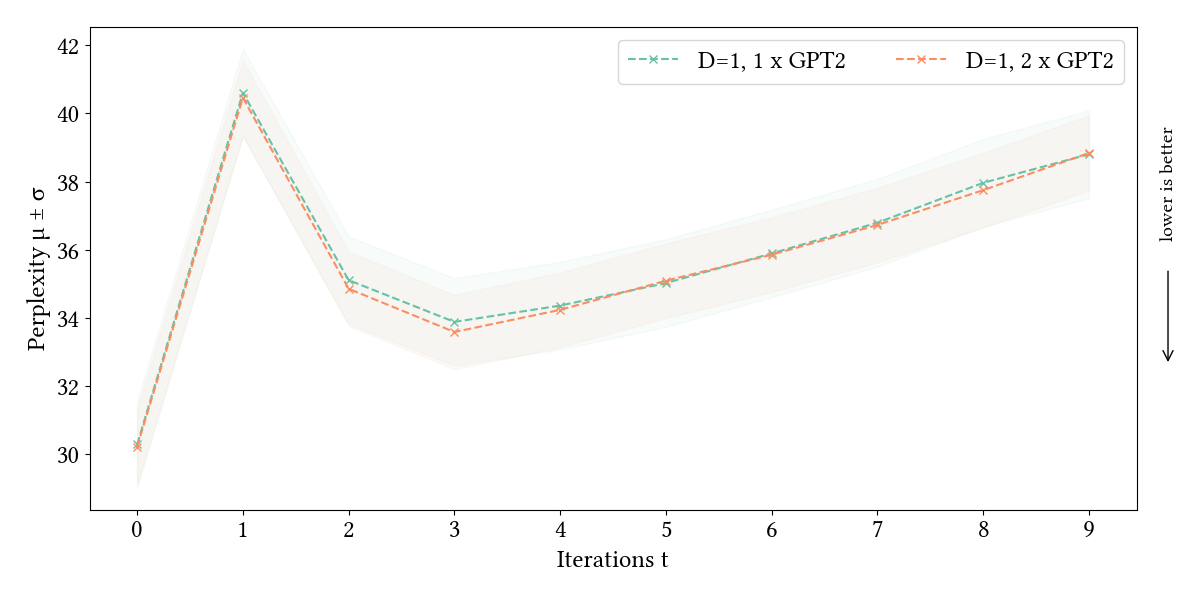}
        \caption{GPT-2}
        \label{fig:gpt2_control}
    \end{subfigure}
    \caption{\textbf{Number of models cannot fully explain the performance differences observed across ecosystems.}
        1×OPT-125m and 1×GPT-2: a single model trained on the entire dataset of size $N$, corresponding to the $D=1$ condition in the primary setting (green). 2×OPT-125m and 2×GPT-2: two identical models, each trained on the entire dataset of size $N$ (orange).
        The generation datasets are identical for all models. The training datasets are identical at the first iteration step but may differ slightly in later iterations for the two-model runs due to stochasticity in setup. The shaded region represents the combined standard deviation across models and data points.}
    \label{fig:control_number_models}
\end{figure}

\begin{figure}[htbp]
    \centering
    \begin{subfigure}[t]{0.5\textwidth}
        \centering
        \includegraphics[width=\linewidth]{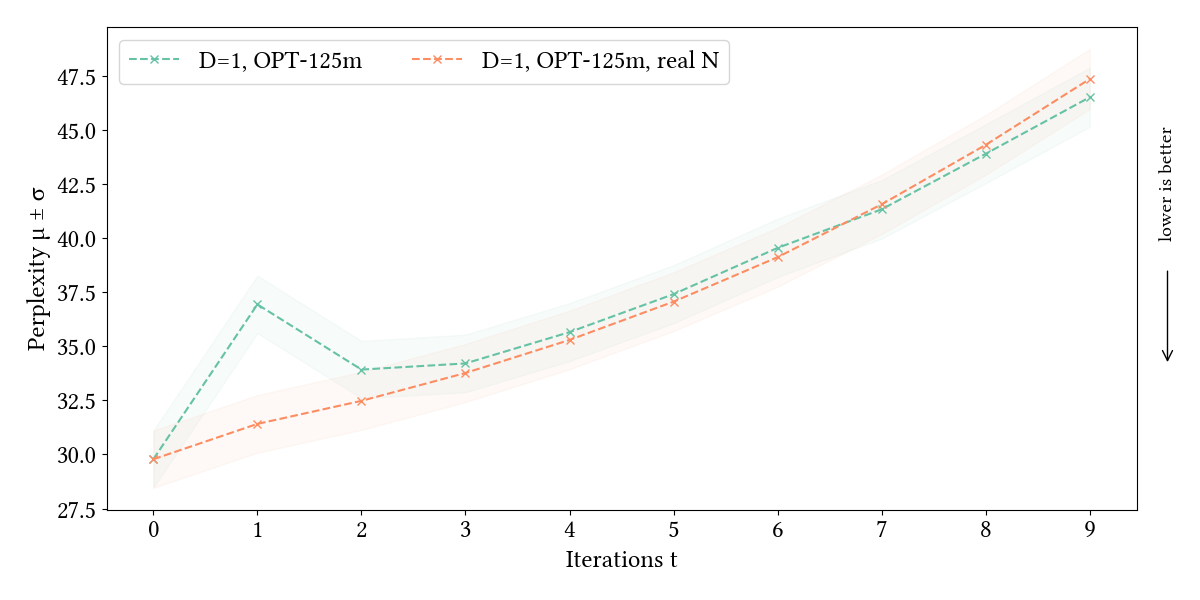}
        \caption{OPT-125m}
        \label{fig:opt_baseline_comp}
    \end{subfigure}
    ~ 
    \begin{subfigure}[t]{0.5\textwidth}
        \centering
        \includegraphics[width=\linewidth]{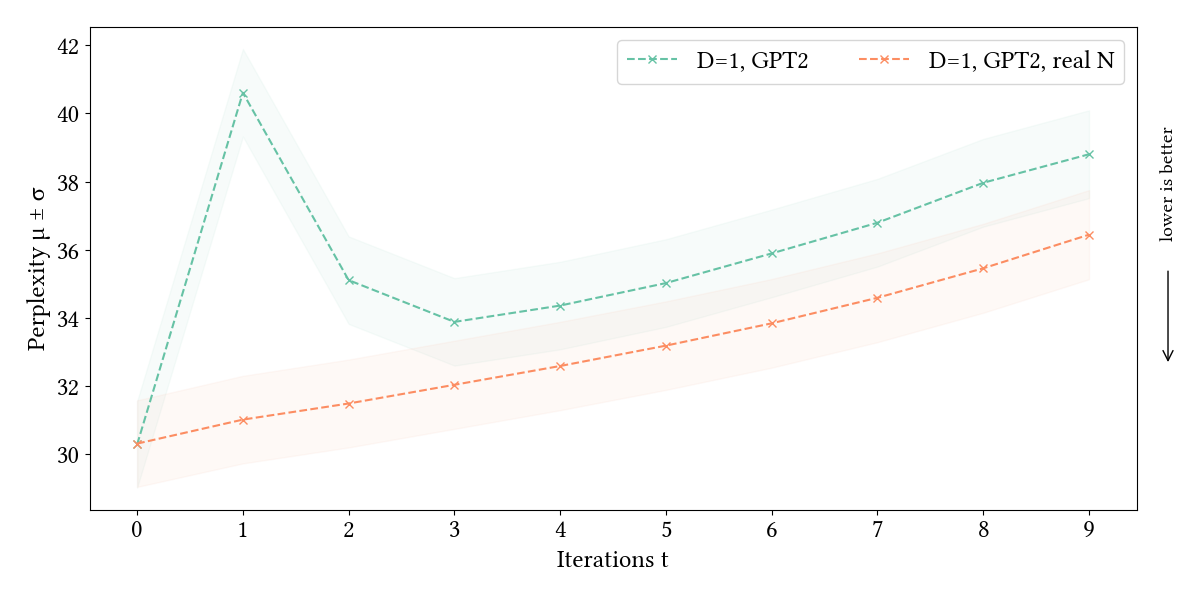}
        \caption{GPT-2}
        \label{fig:gpt2_baseline_comp}
    \end{subfigure}
    \\ 
    \centering
    \begin{subfigure}[t]{0.5\textwidth}
        \centering
        \includegraphics[width=\linewidth]{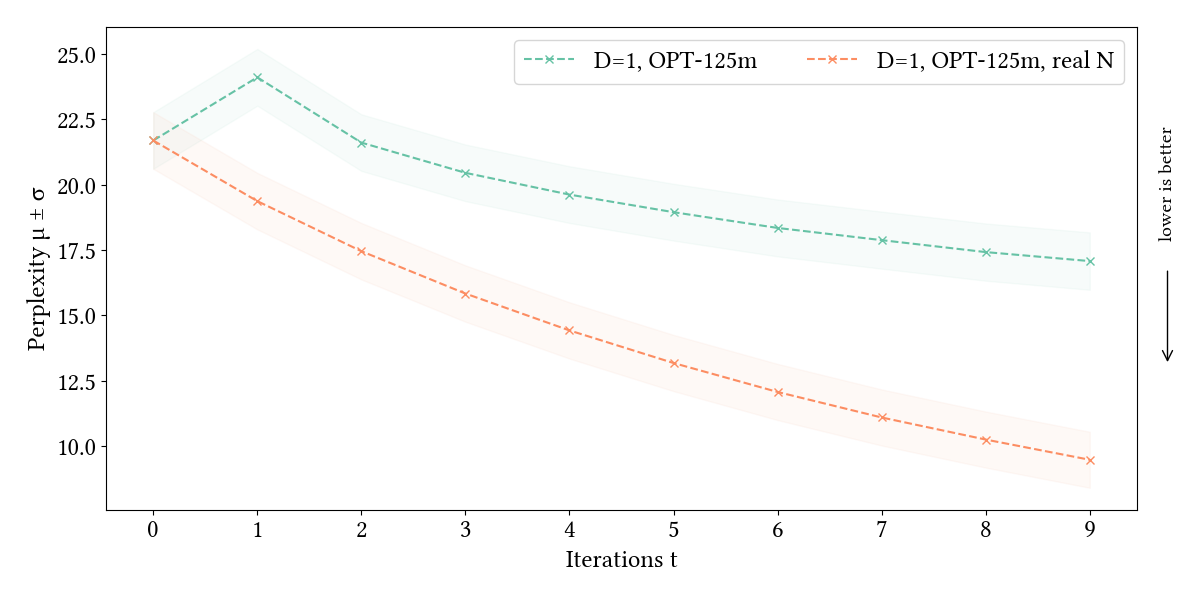}
        \caption{OPT-125m evaluated on train set}
        \label{fig:opt_bl_ontrain_comp}
    \end{subfigure}
    ~ 
    \begin{subfigure}[t]{0.5\textwidth}
        \centering
        \includegraphics[width=\linewidth]{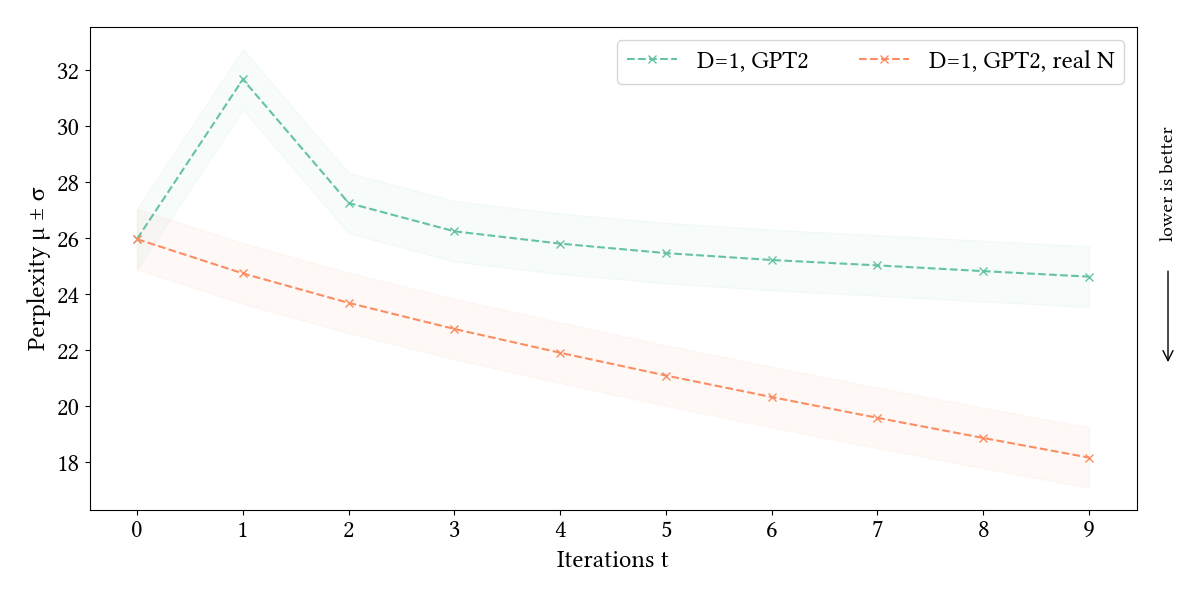}
        \caption{GPT-2 evaluated on train set}
        \label{fig:gpt2_bl_ontrain_comp}
    \end{subfigure}
    \caption{\textbf{Overfitting alone cannot explain the observed performance decay.}
Comparison of the self-training loop in the primary setting (green) with a re-training loop on the fixed real training set (orange), evaluated on the test set (top) and training set (bottom). Because the training data in the re-training condition remains fixed, any performance decay can be fully attributed to overfitting. The lower perplexity on the training set in the re-training condition indicates stronger (over)fitting to the fixed training data. This suggests that overfitting alone cannot account for the performance decline observed in the self-training loop, which appears comparable for OPT-125m and even stronger for GPT-2. }
    \label{fig:control_overfitting}
\end{figure}

We conducted additional experiments to isolate factors that might affect the performance changes observed in the primary setting. The results indicate that neither (model-independent) data quality (Figure~\ref{fig:control_dataquality}), training data size (Figure~\ref{fig:control_datasize}), the number of models per ecosystem (Figure~\ref{fig:control_number_models}), nor overfitting (Figure~\ref{fig:control_overfitting}) can fully account for the observed performance decay or the differences across ecosystems. However, overfitting, which is captured by the EDQ, appears to be an important contributor to performance decay.

Because these individual factors fail to explain the performance trajectories in isolation, the findings provide support for the EDQ hypothesis.

\section{Hyperparameters}
Table~\ref{tab:generation_hyperparameters} provides the  hyperparameters used for text generation.

\begin{table}[h!]
\centering
\begin{tabular}{ll}
\hline
\textbf{Hyperparameter} & \textbf{Value} \\
\hline
temperature      & 0.0 \\
do\_sample & False \\
max\_new\_tokens   & 128 \\
min\_new\_tokens   & 128 \\
repetition\_penalty & 0.0 \\
batch size       & 32 \\
block size       & 128 \\
num\_beams & 5 \\
\hline
\end{tabular}
\caption{Generation hyperparameters.}
\label{tab:generation_hyperparameters}
\end{table}

\end{document}